\newcommand{\ks}[1]{{\color{red}{[#1]}}}

\documentclass[10pt,twocolumn,letterpaper]{article}

\usepackage{cvpr}              % To produce the CAMERA-READY version
\usepackage{wrapfig}
\usepackage{booktabs}
\usepackage[dvipsnames]{xcolor}
\usepackage{soul}
\usepackage{comment}
\usepackage{bbm}
\usepackage{multirow}
\usepackage{arydshln}
\usepackage[linesnumbered,ruled,vlined]{algorithm2e}
\usepackage{adjustbox}

\newcommand{\red}[1]{{\color{red}#1}}
\newcommand{\blue}[1]{{\color{blue}#1}}

\newlength\savewidth\newcommand\shline{\noalign{\global\savewidth\arrayrulewidth\global\arrayrulewidth1.25pt}\hline\noalign{\global\arrayrulewidth\savewidth}}
\newcommand{\std}[1]{\footnotesize{#1}}

\usepackage{amsmath,amsfonts,bm}

\def\eqref#1{Eq.~(\ref{#1})}
\def\1{\bm{1}}

\DeclareMathAlphabet{\mathsfit}{\encodingdefault}{\sfdefault}{m}{sl}
\SetMathAlphabet{\mathsfit}{bold}{\encodingdefault}{\sfdefault}{bx}{n}

\def\gB{{\mathcal{B}}}
\def\gC{{\mathcal{C}}}
\def\gD{{\mathcal{D}}}

\def\gF{{\mathcal{F}}}

\def\gL{{\mathcal{L}}}
\def\gM{{\mathcal{M}}}
\def\gN{{\mathcal{N}}}

\def\gS{{\mathcal{S}}}
\def\gT{{\mathcal{T}}}

\def\sR{{\mathbb{R}}}

\newcommand{\E}{\mathbb{E}}

\DeclareMathOperator*{\argmin}{arg\,min}

\definecolor{cvprblue}{rgb}{0.21,0.49,0.74}
\usepackage[pagebackref,breaklinks,colorlinks,citecolor=cvprblue]{hyperref}
\usepackage{multirow}

\def\paperID{*****} % *** Enter the Paper ID here
\def\confName{CVPR}
\def\confYear{2026}

\title{Memory-efficient Continual Learning with Prototypical Exemplar Condensation}
       
\author{M.-Duong Nguyen$^{1,3}$, Thien-Thanh Dao$^{4}$, Le-Tuan Nguyen$^{1,3}$, Dung D. Le$^{1,2,3}$, Kok-Seng Wong$^{1,2}$
\\
College of Engineering and Computer Science, VinUniversity$^{1}$, Hanoi, Vietnam\\
Center for Environmental Intelligence, VinUniversity$^{2}$, Hanoi, Vietnam\\
Center for AI Research, VinUniversity$^{3}$, Hanoi, Vietnam\\
Faculty of Information Systems, Phenikaa University$^{4}$, Hanoi, Vietnam\\
{\tt\small duong.nm2@vinuni.edu.vn, thanh.daothien@phenikaa-uni.edu.vn,} \\ {\tt \small\{tuan.nl, dung.ld, wong.ks\}@vinuni.edu.vn}
}

\begin{document}
\maketitle

\begin{abstract}
% The ABSTRACT is to be in fully justified italicized text, at the top of the left-hand column, below the author and affiliation information.
% Use the word ``Abstract'' as the title, in 12-point Times, boldface type, centered relative to the column, initially capitalized.
% The abstract is to be in 10-point, single-spaced type.
% Leave two blank lines after the Abstract, then begin the main text.
% Look at previous \confName abstracts to get a feel for style and length.

Rehearsal-based continual learning (CL) mitigates catastrophic forgetting by maintaining a subset of samples from previous tasks for replay. Existing studies primarily focus on optimizing memory storage through coreset selection strategies. While these methods are effective, they typically require storing a substantial number of samples per class (SPC), often exceeding 20, to maintain satisfactory performance. In this work, we propose to further compress the memory footprint by synthesizing and storing prototypical exemplars, which can form representative prototypes when passed through the feature extractor. Owing to their representative nature, these exemplars enable the model to retain previous knowledge using only a small number of samples while preserving privacy. Moreover, we introduce a perturbation-based augmentation mechanism that generates synthetic variants of previous data during training, thereby enhancing CL performance.
Extensive evaluations on widely used benchmark datasets and settings demonstrate that the proposed algorithm achieves superior performance compared to existing baselines, particularly in scenarios involving large-scale datasets and a high number of tasks.
\end{abstract}    
\section{Introduction}
\label{sec:intro}
In recent years, deep learning has advanced significantly, driving the need for continual learning (CL). In CL, data arrive sequentially over time, requiring deep models to learn and adapt dynamically to evolving information. However, this continual adaptation introduces a critical challenge known as catastrophic forgetting \cite{lange2021continual}, wherein the integration of new knowledge interferes with previously acquired knowledge, often leading to substantial performance degradation. While addressing catastrophic forgetting is already challenging, the problem becomes even more complex in real-world applications, such as autonomous systems \cite{nguyen2024tinympc, yang2025human}, edge devices \cite{cheng2024survey, lv2024ptq4sam}, and personalized services \cite{chen2024mixed}, where models must operate on resource-constrained devices with limited computational power and memory. Furthermore, as the number of tasks continues to grow \cite{zhou2024class}, the demand for efficient and scalable CL methods becomes critical. These constraints highlight the need for approaches that can both mitigate catastrophic forgetting and maintain performance at scale, even under strict resource limitations.

Various CL methods have been proposed to mitigate catastrophic forgetting \cite{asadi2023prototype, wei2023online, bang2022online, bang2021rainbow, yang2023efficient, zhou2025ferret, ye2025online, yoo2024layerwise, yoon2022online, tong2025coreset, hao2023bilevel}. 
% Among them, replay-based methods have shown promising performance by storing a subset of data from old classes as exemplars for experience replay. 
One widely adopted strategy to mitigate \textbf{catastrophic forgetting} is \textbf{rehearsal-based approaches}~\cite{deng2025unlocking, wang2025cut, hacohen2025predicting, zheng2024multilayer, urettini2025online}. 
Fundamental, the rehearsal-based approaches can be categorized into three sub-categories, i.e., experience replay, generative replay, and gradient episodic memory. 
Experience replay retains a small subset of samples from previously learned tasks and reuses them during training to alleviate catastrophic forgetting.
Generative replay \cite{shin2017continual} replaces stored samples with a generative model that synthesizes pseudo-samples of past tasks, thereby reducing memory requirements while still enabling rehearsal. However, generative replay may still inherit the problem of catastrophic forgetting, since the generative model itself must be continually updated.
Gradient episodic memory (GEM) \cite{lopez2017gradient} constrains gradient updates on new tasks by leveraging stored samples, ensuring that the new learning process does not increase the loss on previously seen tasks. Nevertheless, GEM often requires substantial memory, as the dimensionality of stored gradients is comparable to that of the model parameters, which is typically much higher than the dimensionality of the raw data.
In this work, we focus on approaches based on experience replay.

In experience replay methods, it is crucial to maintain a compact yet representative set of exemplars that effectively capture the underlying class distributions of the dataset. Recent studies have explored coreset selection \cite{asadi2023prototype, wei2023online, bang2022online, bang2021rainbow, yang2023efficient, zhou2025ferret, ye2025online, yoo2024layerwise, yoon2022online, tong2025coreset, hao2023bilevel}, which aim to select a subset of samples that maximizes the retention of task-relevant knowledge. While these approaches enhance the quality of rehearsal data, they generally require storing a relatively large number of exemplars per class (see Section~\ref{sec:motivation}). In real-world CL scenarios, as the number of tasks grows \cite{cheng2024survey}, the corresponding increase in the number of classes leads to significant memory overhead. Consequently, these methods impose substantial memory demands, thereby limiting their scalability in practice.

Acknowledging these limitations, we introduce Prototypical Coreset Condensation (ProtoCore), a prototypical exemplar condensation framework for CL. ProtoCore directly learns a set of exemplars that closely approximate the class prototypes, thereby maintaining consistency with both the current task and the prototypical exemplars from previous tasks. By adopting this direct learning strategy, ProtoCore mitigates the representation shift caused by reliance on a feature extractor (see Appendix~\ref{sec:motivation}), which is a key contributor to catastrophic forgetting in CL.
Our main contributions are as follows:
\begin{enumerate}
    \item We address the representation shift introduced by relying on a feature extractor by proposing a method to learn synthetic exemplars whose embeddings closely approximate their corresponding class prototypes.
    \item To improve the quality of the learned prototypical synthetic exemplars, we adopt a prototypical network as the feature extractor and introduce a prototypical alignment loss. This loss explicitly guides the network to preserve previously learned knowledge while mitigating catastrophic forgetting, relying solely on stored prototypical exemplars instead of the original training data.
    \item To maximize memory efficiency with only a few samples per class, we design a surrogate continual learning strategy for training the classifier. This strategy uses a prototype perturbation technique \cite{zhu2021prototype} to generate synthetic samples directly from learned prototypes, enabling substantial mitigation of catastrophic forgetting while requiring only a single exemplar per class in memory.
    \item We evaluate our proposed methods against recent baselines across a range of challenging datasets. The experimental results demonstrate that our approach achieves performance competitive with state-of-the-art baselines, while offering substantial improvements in memory.
    \item We also investigate long task sequences and observe that many existing methods are not effective at mitigating catastrophic forgetting in this regime. ProtoCore demonstrates strong performance under long sequence settings, indicating promising directions for future research.
\end{enumerate}
\begin{comment}
This fixed and minimal replay requirement demonstrates particular efficiency when the number of tasks grows large. In conventional replay-based methods, the number of stored samples per class typically increases proportionally with the number of tasks \cite{}, whereas in ProtoCore, this number remains constant at one sample per class, offering a scalable and memory-efficient solution.

\hl{Memory Cost, Online + Incremental, others?}
\begin{itemize}
    \item Only requires 1 sample per class (SPC).
    \item Can be applied on online + incremental + ...
\end{itemize}
\end{comment}

% \subsection*{Notations}

\section{Background \& Preliminaries}
\label{sec:preliminaries}
This section offers a primer on online CL, prototypical networks, and dataset distillation. Particular emphasis is placed on the online formulation of prototypical networks and the configuration of dataset condensation within a CL framework, as these elements are critical to our method exposition in Section~\ref{sec:method}.
\subsection{Online Continual Learning}
Consider a sequence of tasks $\gT = \{\gT_1, \gT_2, \gT_3, \ldots, \gT_T\}$, each of which consists of unique data and classes $\gT_t = \{(x_{t,i}, y_{t,i})\}^{\vert\gT_t\vert}_{i=1}$, where $T$ is the total task number and $\vert\gT_t\vert$ is the number of data according to task $\gT_t$. At every task $\gT_t$, there are subset $\gC_t$ of $\vert C_t\vert$ classes available, and we have $\gC = \gC_{1:T} = \{\gC_1 \bigcup \gC_2 \bigcup \ldots \gC_T\}$.

Yet, for the online setting, each data sample inside the task $(x_{t,i}, y_{t,i})$ is passed only once in a non-stationary stream, with no task identity provided. The target of CL is to maximize the overall accuracy of all the tasks.

Replay-based methods maintain a small auxiliary memory $\gM_x = \{(\hat{x}_i, \hat{y}_i)\}^{\vert M_x\vert}_{i=1}$ to store images from past tasks, where $\vert\gM_{x}\vert$ is the memory size. 
During each iteration of current task, a batch of samples $\gB_m = \{(\hat{x}_{m,i}, \hat{y}_{m,i})\}^{\vert \gB_m\vert}_{i=1}$ is sampled from the memory randomly or under certain strategies for joint training with current data $\gB_t=\{(x_{t,i}, y_{t,i})\}^{\gB_t}_{i=1}$, where $\vert\gB_m\vert < \vert\gM_x\vert$ and $\vert\gB_t\vert$ are the mini-batch size of the replay data and the current tasks, respectively. We denote $\vert\cdot\vert$ as both of the absolute value of a scalar, and the size of a set. The joint training process is formulated as follows: 
\begin{align}
    \phi^* = \argmin_{\phi} \gL_\gT(\theta,\phi;\gT_t) + \lambda\gL_\gM(\theta,\phi;\gM_x),
\end{align}
where $\theta,\phi$ are the parameters of the feature extractor and classifier, respectively. $\gL_\gT(\cdot;\cdot)$ is the standard training objective for the task, and $\gL_\gM(\cdot;\cdot)$ is the objective for the replay, and $\lambda$ is a coefficient hyper-parameter for the replay objective. By simultaneously viewing the data from the current and past tasks, the model can obtain new knowledge while reduce the catastrophic forgetting for the past. However, when the storage space is restricted, the memory informativeness become critical for effective replay. And we argue that by only employing the original images, the storage space is still under-utilized. 

\subsection{Prototypical Networks}
Prototypical Networks (PNs)~\cite{snell2017prototypical} are a class of metric-based meta-learning algorithms designed to generalize to new classes with limited labeled data. 
Instead of directly learning a parametric classifier, PNs learn a metric space in which classification is performed by computing distances between query samples and class prototypes. 

Given a set of classes $\mathcal{C}$, each class $c \in \mathcal{C}$ is represented by a prototype vector, computed as the mean embedding of its support set:
\begin{equation}
    \mathbf{p}_c = \frac{1}{|S_c|} \sum_{\mathbf{x}_i \in S_c} f_\theta(\mathbf{x}_i),
\end{equation}
where $S_c$ is the support set for class $c$, $f_\theta(\cdot)$ is the feature encoder parameterized by $\theta$, and $\mathbf{p}_c \in \mathbb{R}^d$ denotes the class prototype in the embedding space.

For a query sample $\mathbf{x}_q$, classification is performed by computing its distance to each class prototype using a metric function $d(\cdot,\cdot)$, typically the squared Euclidean distance. 
The probability that $\mathbf{x}_q$ belongs to class $c$ is given by:
\begin{equation}
    P(y = c \mid \mathbf{x}_q) = 
    \frac{\exp(-d(f_\theta(\mathbf{x}_q), \mathbf{p}_c))}
    {\sum_{c' \in \mathcal{C}} \exp(-d(f_\theta(\mathbf{x}_q), \mathbf{p}_{c'}))}.
\end{equation}
The loss encourages the encoder to produce embeddings close to their class prototype while being far from others.

\noindent\textbf{Relation to Contrastive Learning.}
Prototypical networks share a strong conceptual connection with contrastive learning, which learns representations by pulling similar samples together and pushing dissimilar samples apart in the embedding space.
A typical objective for contrastive learning is the InfoNCE loss~\cite{oord2018representation}:
\begin{equation}
    \mathcal{L}_{\mathtt{contrastive}} = 
    - \log \frac{\exp(\mathrm{sim}(f_\theta(\mathbf{x}_i), f_\theta(\mathbf{x}_j))/\tau)}
    {\sum_{k} \exp(\mathrm{sim}(f_\theta(\mathbf{x}_i), f_\theta(\mathbf{x}_k))/\tau)},
\end{equation}
where $\mathrm{sim}(\cdot,\cdot)$ is a similarity function (e.g., cosine similarity), and $\tau$ is a temperature parameter.

From this perspective, the prototypical network loss can be viewed as a class-level contrastive objective. 
Instead of contrasting individual instance pairs, PNs contrast query samples against class prototypes, which serve as representative anchors for each class:
\begin{itemize}
    \item \textbf{Positive pair:} the query sample and its true class prototype.
    \item \textbf{Negative pairs:} the query sample and all other class prototypes.
\end{itemize}

This equivalence highlights that prototypical networks can be interpreted as a structured form of contrastive learning, where the prototype acts as an aggregated representation that reduces noise and provides better generalization. 
This property makes PNs particularly effective in few-shot and online CL scenarios, where data arrive incrementally and efficient representation learning is crucial.

\begin{figure*}[!ht]
\centering
\includegraphics[width = \linewidth]{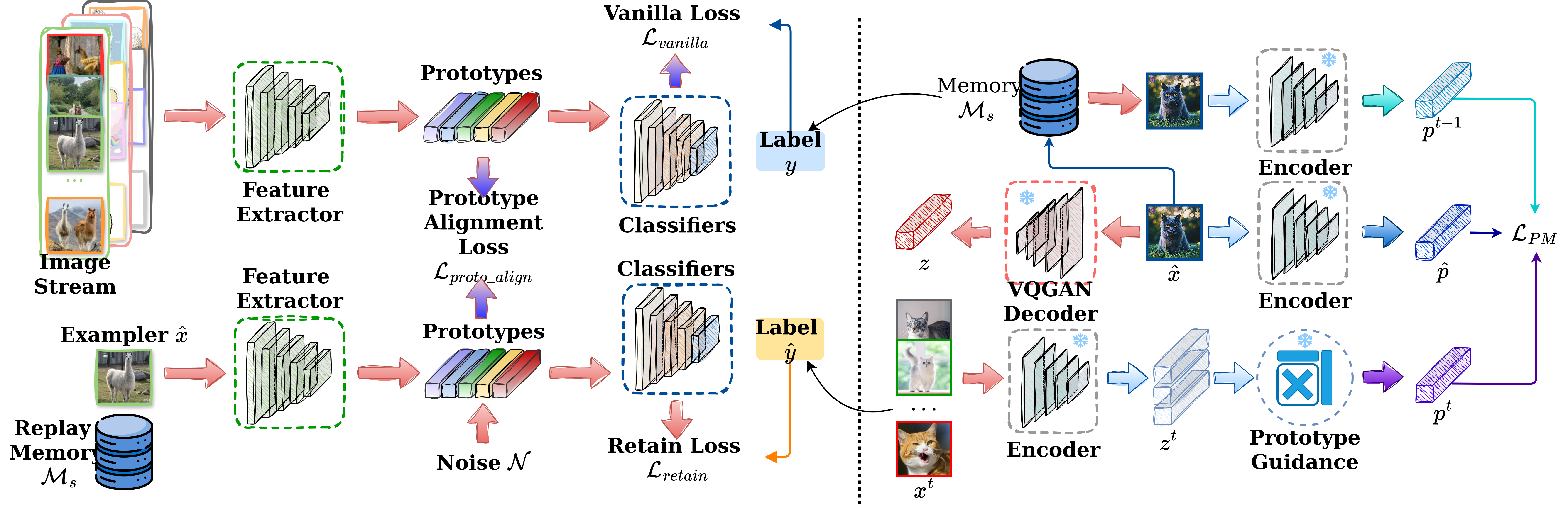}
\caption{Illustration of ProtoCore architecture.}
\label{fig:ProtoCore}
\end{figure*}

\subsection{Dataset Distillation}
In the concept of CL, the goal of dataset condensation (DC) is at every task $t$, we distill a large dataset $\gT_t=\{(x_i, y_i)\}^{\vert\gT_t\vert}_{i=1}$ containing $\vert\gT_t\vert$ training images $x_i\in\sR^{d}$ and its labels $y_i\in\{1,2,\ldots,\vert Y\vert\}$ into a small dataset $\gS_t = \{s_i, y_i\}^{\vert\gS_t\vert}_{i=1}$ \cite{yang2023efficient} with $\vert \gS_t\vert$ synthetic image $s_i\in\sR^{d}$, where $\vert \gS_t\vert \ll \vert \gT_t\vert$ (2-3 orders of magnitude), $\vert Y\vert$ represents the number of classes, and $\sR^{d}$ defines a $d$-dimensional space. We expect a network $\phi_{\theta^{\gT}}$ trained on the large training set $\gT$ on the unseen test dataset, that is: 
\begin{align}
    &\E_{x_i\sim P_\gT}[\ell (\phi_{\theta^{\gT}}(x_i), y)] 
     \simeq
     \E_{x_i\sim P_\gS}[\ell (\phi_{\theta^{\gT}}(x_i), y)], \\
    \textrm{s.t.}
    &~~\theta^{\gT} = \argmin_{\theta^{\gT}} \gL^{\gT}(\theta^{\gT}) 
                   = 
                   % \argmin_{\theta^{\gT}} 
                   \frac{1}{\gN_\gT}\sum_{(x,y)\in\gT} \ell (\phi_{\theta^{\gT}}(x_i), y) \notag \\
    &~~\theta^{\gS} = \argmin_{\theta^{\gS}} \gL^{\gS}(\theta^{\gS}) 
                   = 
                   % \argmin_{\theta^{\gS}} 
                   \frac{1}{\gN_\gS}\sum_{(x,y)\in\gS} \ell (\phi_{\theta^{\gS}}(x_i), y) \notag
\end{align}
where $P_\gT$ represents the real distribution of the test dataset, $x_i$ represents the input image, $y$ represents the ground truth, and $\ell(\cdot)$ represents a loss function such as cross-entropy.

\section{Method}\label{sec:method}
In this section, we present our proposed \textbf{ProtoCore} framework. 
Overall, ProtoCore comprises three main components: 
(1) an efficient replay memory that stores prototypical exemplars, 
(2) prototypical exemplar generation via model inversion, and 
(3) CL with prototypical exemplars through prototype perturbations. 
An overview of the framework is illustrated in Figure~\ref{fig:ProtoCore}.\subsection{Efficient Replay Memory}
To design the efficient replay memory which only contains the prototypical exemplars, in which contain only one sample per class, we design a prototypical exemplar generation which based on the model inversion technique \cite{wei2025open}. Specifically, we inverse the knowledge from the model trained on the current task to generate exemplars that serve as prototypes for each class. As a consequence, we can ignore the utilization of the generative models in previous works \cite{ho2023prototype, zhu2021prototype}. It is worth noting that the use of generative models with dimensionality $M$ introduces considerable computational overhead, as backpropagation must be performed through the generative network, in contrast to directly generating exemplars of size $C \times H \times W$ from model inversion. 

\subsection{Prototypical Exemplar Generation.}
Our objectives are threefold. First, the synthetic exemplars should maintain minimal distance to the learned prototypes of the current tasks, ensuring they accurately represent the newly acquired knowledge. Second, they should also remain close to the prototypes of the previous tasks, thereby preserving prior knowledge. Third, we should avoid the representation shift due to the synthetic data generation \cite{hu2025task}. Consequently, the joint optimization is as follows:
\begin{align}
    z = \argmin_{z} \gL_\mathtt{cur\_syn} + \gL_\mathtt{pre\_syn} + \gL_\mathtt{shift}. 
\end{align}
Given the optimized latent variable $z$, the corresponding synthetic image is produced by the pretrained decoder as $s = g(z)$. This joint optimization significantly mitigates catastrophic forgetting, as the model is encouraged to selectively retain the most salient and task-relevant features from past tasks while discarding trivial information. As a result, the learned memory remains stable and resistant to forgetting.
\subsubsection{Current Exemplar Alignment.}
To align the exemplars with the prototypical representations of the current task, the core idea is to synthesize data that produces feature representations closely matching the prototypes of the current task. 
Given a target class $t$, the prototypical exemplars can be synthesized using the feature extractor $f_{\theta}$ as follows:
\begin{align}
    \gL_\mathtt{cur\_syn} = \sum^{}_{c\in\gC_t}d(f_{\theta}(s^c); p^c),
\end{align}
where $p^c = \frac{1}{\vert \gD^c_t\vert}\sum_{(x;\cdot)\sim\gD^c_t}f_{\theta}(x)$ denotes the prototype of class $c$ computed from the real data of the current task $t$.
Using the feature extractor, the synthetic exemplar $s^c$ is optimized to form synthetic prototypes that approximate the real ones.
The function $d$ measures the distance between the synthetic and real prototypes. In our work, we adopt the mean squared error (MSE) as the distance metric.

% Edit synthetic image $s$ by reducing the distance between target prototypes and the representations of synthetic exemplars.

\subsubsection{Previous Exemplar Alignment.}
One distinctive characteristic of ProtoCore is its ability to continuously align the current prototypes with the synthetic prototypes derived from previous tasks. Unlike prior works \cite{ho2023prototype, zhu2021prototype, lin2022prototype}, which perform prototype aggregation through simple or soft averaging between current and previously stored prototypes, our approach focuses on constructing prototypes via exemplar synthesis. This design enables the synthesized exemplars to preserve as much discriminative information as possible from both the previously stored prototypes and the current task representations.
Formally, the alignment loss with respect to the previous task is defined as follows:
\begin{align}
    \gL_\mathtt{pre\_syn} = \sum_{c\in \gC_{1:t}} d(f_{\theta}(s^c); f_{\theta}(\hat{s}^c)),
\end{align}
where $\hat{s}^c$ are the stored synthetic exemplars that capture the information of previous tasks and drawn from $\gM_s$.

\noindent\textbf{Discussion on synthesize over misclassified samples.} Since the prototypes are computed from real data, the feature extractor may generate representations in which some samples are not well captured. Consequently, the prototype estimation 
$p^c = \frac{1}{|\gD^c_t|}\sum_{(x, \cdot) \sim \gD^c_t} f_{\theta}(x)$
can become inaccurate when computed using mini-batches. To mitigate this issue, we refine the prototype computation by excluding samples that are misclassified by the classifier. This filtering process allows us to estimate more reliable class prototypes, thereby improving the effectiveness of the prototypical exemplar alignment and leading to better overall performance.

\noindent\textbf{Discussion on Representation Shift.} Similar to the phenomenon observed with synthetic data in meta-learning \cite{hu2025task}, the evolving nature of synthetic datasets can induce shifts in the distribution of prototypical exemplars stored in memory. This effect becomes more pronounced as the number of tasks increases and the synthetic data generation process is repeated multiple times. To maintain consistency in CL when synthetic exemplars are employed, we store the real prototypes $\hat{p}^c$ for each class in memory. These real prototypes serve as anchors that guide the synthetic exemplars, preventing them from drifting excessively from the true representations. Specifically, we introduce the following regularization term:
\begin{align}
\mathcal{L}_\mathtt{shift} = \sum_{c \in \gC_{1:t}} d(f_{\theta}(s^c), \hat{p}^c),
\end{align}
Since only one real prototype is retained per class, the additional memory overhead introduced by this mechanism remains minimal.

% \paragraph{Discussion on latent initialization.}

\subsection{Continual Learning with Prototypical Exemplars}
In this work, our objective is to develop a memory-efficient CL approach that requires storing only few sample per class. This constraint fundamentally distinguishes our method from existing approaches, as we cannot directly use exemplars from the replay memory to mitigate catastrophic forgetting from previous tasks. The limitation arises because a single sample per class is insufficient to achieve strong class discrimination, potentially leading to imbalanced learning when combined with data from current tasks. To address this challenge, we aim to leverage class prototypes generated from the exemplars and synthesize additional data through controlled perturbations, following the strategy proposed in \citet{zhu2021prototype}.
\subsubsection{Incremental Feature Extractor}
To enhance the quality of synthetic prototypical exemplars, it is essential to construct accurate real prototypes for each task. Consequently, we integrate a prototypical network into the feature extractor. Within the CL framework, the objective of the prototypical network is to learn representations that effectively capture both current and previously learned tasks. To achieve this, we introduce two prototypical loss functions: one for the current task and another for the previously learned tasks.
If class $c$ is available in classes of current task, we deploy the prototypical loss on the current data task as follows:
\begin{align}
    \gL^c_\mathtt{cur\_pro} = \sum_{(x^c,y^c) \in \gD_t}\frac{\exp{(-d(f_{\theta}(x^c);p^c))}}{\sum_{c{'}\neq c}\exp{(-d(f_{\theta}(x^c);p^{c{'}}))}},
\end{align}
where $d$ represents the statistical loss (e.g., MSE) between two representations.  
To mitigate the forgetting issue in the feature extractor as tasks progress, we design a prototypical mechanism that aligns the current prototypes with the stored prototypical exemplars. Specifically, when class $c$ is available in the replay memory, we leverage its exemplars to approximate the distribution of previously seen classes and compute the prototypical loss on prior tasks accordingly.
\begin{align}
    \gL^c_\mathtt{pre\_pro} = \frac{\exp{(-D(\mathbf{z}^c;p^c))}}{\sum_{c{'}\neq c}\exp{(-D(\mathbf{z}^c;p^{c{'}}))}},
\end{align}
where $\mathbf{z}^c = \{ f_{\theta}(h(s^c)) \mid \forall h \in \mathcal{F} \}$ denotes the set of feature representations generated by applying perturbations $h(\cdot)$ from the transformation set $\mathcal{F}$ to the synthetic exemplars $s^c$ of class $c$.
\newline
\textbf{Discussion on Prototypical Loss.} We initially experimented with the prototypical loss \cite{snell2017prototypical}. However, the resulting class representations were not well-separated. Well-separated classes are crucial for learning high-quality prototypes, as they enable the prototypes to more accurately represent their respective classes and allow perturbations among prototypes to better capture class-level variations. Consequently, we opted to use contrastive loss instead. Contrastive loss explicitly encourages separation between samples from different classes, leading to more distinct and informative prototypes when computed as the class centers.
% \paragraph{Discussion on Adaptive Temperature.}

\begin{comment}
\begin{align}
    \gL_{pre\_syn} = \sum_{c\in C_{1:t}} d(\hat{p}^{c}; f_{\theta}(\hat{s}^c)),
\end{align}
where $\hat{p}^{c}$ is the aligned prototype according to class $c$ and can be computed as follows: 
\begin{align}
    \hat{p}^{c} 
    = &\frac{\beta_1}{\vert \gM^c_x\vert}\sum_{(\hat{x};\cdot)\sim\gM^c_x}f_{\theta}(\hat{x})
    + \frac{\beta_2}{\vert \gD^c_t\vert}\sum_{(x;\cdot)\sim\gD^c_t}f_{\theta}(x) \notag \\
    + &~\frac{\beta_3}{\vert\gF\vert}\sum_{h\sim\gF}f_{\theta}(h(\hat{s}^c)).
\end{align}
Samples $\gM^c_x$ according to classes $c$ in $\gM_x$.
Samples $\gD^c_t$ according to classes $c$ in $\gD_t$.
\end{comment}

\subsubsection{Task-head Learning with Memory}
To alleviate catastrophic forgetting in CL settings, we apply the task loss to both the current data and the stored exemplars from previous tasks. For the current task, standard task learning is performed using only the classes available in the current dataset. For instance,
\begin{align}
    \mathcal{L}_{\mathtt{task\_cur}} = \mathcal{L}_{\mathtt{task}}(f_{\phi}(z), y),
\end{align}
where the latent feature $z$ is obtained from the feature extractor $f_{\theta}$ as $z = f_{\theta}(x)$ for all $x \in \mathcal{D}_t$. 
The key distinction between ProtoCore and other CL approaches lies in the use of class-specific prototypical exemplars $s^c$, which help the model retain knowledge from previous tasks. However, since only a limited number of SPCs are stored, it is often challenging for existing methods to effectively perform rehearsal on past tasks. 
To address this limitation, our method adopts the prototype augmentation technique proposed by \cite{zhu2021prototype} to generate diverse synthetic latent representations for each class $c$. Specifically,
\begin{align}
    \mathcal{L}_{\mathtt{task\_pre}} = \sum_{c\in \gC_{1:t}}\mathcal{L}^c_{\mathtt{task}}(f_{\phi}(\mathbf{z}^c), c),
\end{align}
where $\mathbf{z}^c = \{ f_{\theta}(s^c) + \zeta \mid \zeta \sim \mathcal{N}(0, \mathbb{I}\sigma^2) \}$ and $\zeta$ denotes Gaussian noise with the same dimensionality as the prototype. 
By introducing prototype augmentation, we can approximate the underlying data distribution of each class, enabling the model to preserve knowledge of previous tasks even in the absence of real data.
The overall task head learning is calculated as: 
\begin{align}
    \gL_{\mathtt{task}} = \mathcal{L}_{\mathtt{task\_cur}} + \mathcal{L}_{\mathtt{task\_pre}}. 
\end{align}

% \paragraph{Task Loss on previous task.}
% else, we leverage the exemplars to make the distribution of unseen classes.
% \begin{align}
%    \gL_{CE}~+= \gL_{CE}(f_{\phi}(\mathbf{z}), y)\quad \textrm{s.t.}\quad \mathbf{z} = \{f_{\theta}(h(s)) \vert \forall h\in\gF\}
% \end{align}

% \subsection{Compatible with other experience replay memory.}

% \subsection{Problem Definitions}

% \input{CVPR2025/sec/4_theoretical}
% \input{CVPR2025/sec/5_settings}
\begin{table*}[!h]
    \caption{Last ($A_T$) and learning ($A_L$) accuracy (\%) comparison on benchmarks with total number of tasks $T$. Results are averaged over 10 runs with mean $\pm$ standard deviation. \red{\textbf{Best}} and \blue{\textbf{Second Best}} results are highlighted. Our ProtoCore with $20+1$ buffer SPCs indicates that the replay buffer is used to store 20 real samples to support the proposed ProtoCore method.}
    \centering
    \setlength{\tabcolsep}{3pt}
    \begin{adjustbox}{max width=\textwidth}
    {%
    \renewcommand{\arraystretch}{1.3}
    \begin{tabular}{c | c | c | c c | c c | c c | c c}
    \hline
    \multirow{2}{*}{Type} & \multirow{2}{*}{Method} & \multirow{2}{*}{Buffer SPCs} & \multicolumn{2}{c|}{S-CIFAR-100~\cite{dohare2024loss}} & \multicolumn{2}{c|}{S-CIFAR-100~\cite{dohare2024loss}} & \multicolumn{2}{c|}{S-TinyImageNet~\cite{hou2019learning}} & \multicolumn{2}{c}{S-ImageNet-1K~\cite{dohare2024loss}} \\
    & & & \multicolumn{2}{c|}{$T$=10} & \multicolumn{2}{c|}{$T$=50} & \multicolumn{2}{c|}{$T$=20} & \multicolumn{2}{c}{$T$=100} \\
    \cline{4-11}
    & & & $A_T$ & $A_L$ & $A_T$ & $A_L$ & $A_T$ & $A_L$ & $A_T$ & $A_L$ \\
    \hline
    \multirow{5}{*}{\shortstack{Efficient\\Memory}} 
    & CSReL & 20 & $22.71{\scriptstyle\pm0.54}$ & $77.93 {\scriptstyle\pm0.27}$ & $12.78{\scriptstyle\pm0.69}$ & $75.94{\scriptstyle\pm0.33}$ & $13.12{\scriptstyle\pm0.82}$ & $57.44{\scriptstyle\pm0.90}$ & $4.25{\scriptstyle\pm0.48}$ & $56.71{\scriptstyle\pm1.21}$ \\
    & BCSR & 20 & $31.52{\scriptstyle\pm0.12}$ & \blue{$\textbf{78.47}{\scriptstyle\pm0.14}$} & $13.42{\scriptstyle\pm0.24}$ & \blue{$\textbf{80.64}{\scriptstyle\pm0.82}$} & $16.84{\scriptstyle\pm0.51}$ & $55.09{\scriptstyle\pm1.15}$ & $5.05{\scriptstyle\pm0.88}$ & $57.85{\scriptstyle\pm1.44}$ \\
    & OCS & 20 & $29.94{\scriptstyle\pm1.10}$ & $75.89{\scriptstyle\pm0.91}$ & $12.52{\scriptstyle\pm0.66}$ & $77.15{\scriptstyle\pm1.72}$ & $12.87{\scriptstyle\pm1.14}$ & $59.75{\scriptstyle\pm0.77}$ & $4.34{\scriptstyle\pm0.34}$ & $58.92{\scriptstyle\pm0.84}$ \\
    & PBCS & 20 & \blue{$\textbf{35.95}{\scriptstyle\pm1.21}$} & \red{$\textbf{85.34}{\scriptstyle\pm2.13}$} & $14.57{\scriptstyle\pm0.68}$ & \red{$\textbf{86.90}{\scriptstyle\pm0.83}$} & \blue{$\textbf{18.12}{\scriptstyle\pm0.73}$} & \red{$\textbf{74.63}{\scriptstyle\pm1.51}$} & $6.01{\scriptstyle\pm0.41}$ & \red{$\textbf{76.45}{\scriptstyle\pm2.45}$} \\
    & OnPro & 20 & $29.38{\scriptstyle\pm0.94}$ & $77.51{\scriptstyle\pm0.43}$ & $12.64{\scriptstyle\pm0.30}$ & $75.77{\scriptstyle\pm0.27}$ & $14.74{\scriptstyle\pm0.28}$ & $56.58{\scriptstyle\pm1.56}$ & $4.75{\scriptstyle\pm0.44}$ & $58.56{\scriptstyle\pm0.32}$\\
    \hline
    \multirow{4}{*}{\shortstack{Replay\\based}} 
    & iCaRL & 20 & $25.92{\scriptstyle\pm0.78}$ & $55.52 {\scriptstyle\pm0.81}$ & $11.25 {\scriptstyle\pm0.19}$ & $72.52 {\scriptstyle\pm0.37}$ & $10.56 {\scriptstyle\pm0.24}$ & $56.58 {\scriptstyle\pm0.17}$ & $5.05{\scriptstyle\pm0.25}$ & $52.13{\scriptstyle\pm0.41}$ \\
    & DER & 20 & $20.44{\scriptstyle\pm0.72}$ & $71.50{\scriptstyle\pm1.76}$ & $11.63 {\scriptstyle\pm0.10}$ & $73.57 {\scriptstyle\pm0.34}$ & $12.91 {\scriptstyle\pm1.33}$ & $59.24 {\scriptstyle\pm1.16}$ & $4.45{\scriptstyle\pm0.19}$ & $54.93{\scriptstyle\pm0.38}$ \\
    & DER++ & 20 & $23.35{\scriptstyle\pm0.22}$ & $72.26{\scriptstyle\pm0.47}$ & $12.24 {\scriptstyle\pm0.21}$ & $80.51{\scriptstyle\pm1.53}$ & $10.31 {\scriptstyle\pm0.52}$ & $60.35 {\scriptstyle\pm0.74}$ & $4.17{\scriptstyle\pm0.62}$ & $57.84{\scriptstyle\pm1.28}$ \\
    & C-Flat & 20 & $26.81 {\scriptstyle\pm0.94}$ & $65.18{\scriptstyle\pm0.56}$ & $11.84{\scriptstyle\pm0.38}$ & $74.95{\scriptstyle\pm2.05}$ & $14.37{\scriptstyle\pm0.68}$ & $65.98{\scriptstyle\pm1.88}$ & $5.05{\scriptstyle\pm1.17}$ & $63.77{\scriptstyle\pm1.54}$ \\
    \hline
    \multirow{3}{*}{\shortstack{Replay\\free}} 
    & LDC & 0 & $17.54{\scriptstyle\pm0.67}$ & $53.82{\scriptstyle\pm0.76}$ & $9.12{\scriptstyle\pm0.43}$ & $70.12{\scriptstyle\pm1.39}$ & $12.31{\scriptstyle\pm0.44}$ & $50.17{\scriptstyle\pm1.24}$ & $4.82{\scriptstyle\pm0.19}$ & $52.46{\scriptstyle\pm0.71}$ \\
    & FeCAM & 0 & $15.89{\scriptstyle\pm0.24}$ & $50.63{\scriptstyle\pm0.20}$ & $7.26{\scriptstyle\pm0.19}$ & $65.02{\scriptstyle\pm1.82}$ & $12.03{\scriptstyle\pm0.48}$ & $45.98{\scriptstyle\pm1.07}$ & $4.34{\scriptstyle\pm0.22}$ & $47.91{\scriptstyle\pm1.34}$ \\
    & ADC & 0 & $18.68{\scriptstyle\pm0.14}$ & $60.73{\scriptstyle\pm2.10}$ & $11.28{\scriptstyle\pm0.58}$ & $74.56{\scriptstyle\pm2.72}$ & $14.58{\scriptstyle\pm0.58}$ & $65.78{\scriptstyle\pm1.52}$ & $5.97{\scriptstyle\pm0.41}$ & $63.85{\scriptstyle\pm1.48}$ \\
    \hline
    \multirow{3}{*}{Ours} 
    & ProtoCore & 1 & $18.26{\scriptstyle\pm0.48}$ & $75.12{\scriptstyle\pm2.12}$ & $12.31{\scriptstyle\pm1.34}$ & $77.79{\scriptstyle\pm1.59}$ & $14.62{\scriptstyle\pm1.63}$ & $66.15{\scriptstyle\pm2.23}$ & $6.57{\scriptstyle\pm0.34}$ & $64.90{\scriptstyle\pm1.62}$ \\
    
    & ProtoCore & 5 & $31.17{\scriptstyle\pm0.83}$ & $76.85{\scriptstyle\pm1.39}$ & \blue{$\textbf{15.29}{\scriptstyle\pm0.54}$} & $74.98{\scriptstyle\pm2.09}$ & $17.31{\scriptstyle\pm0.37}$ & $67.15{\scriptstyle\pm1.40}$ & \blue{$\textbf{9.03}{\scriptstyle\pm0.22}$} & \blue{$\textbf{67.12}{\scriptstyle\pm1.71}$} \\

    & ProtoCore & 20 + 1 & \red{$\textbf{37.51}{\scriptstyle\pm0.29}$} & $76.58{\scriptstyle\pm2.03}$ & \red{$\textbf{24.11}{\scriptstyle\pm0.44}$} & $76.37{\scriptstyle\pm1.86}$ & \red{$\textbf{27.11}{\scriptstyle\pm0.75}$} & \blue{$\textbf{68.91}{\scriptstyle\pm1.37}$} & \red{$\textbf{14.22}{\scriptstyle\pm0.47}$} & $65.86{\scriptstyle\pm1.81}$ \\
    
    \hline
    \end{tabular}
    }%
    \end{adjustbox}
    \vspace{-0.1cm}
    \label{tab:main_results_with_vlines_and_thick_hlines}
\end{table*}
\section{Experimental Settings}
\textbf{Baselines:}
In this work, we compare our proposed ProtoCore with representative baselines from three major categories: efficient memory for CL, replay-based CL, and replay-free CL. 
These comparisons are designed to address the following key questions: 
(1) Does ProtoCore provides superior memory efficiency and better mitigation of catastrophic forgetting compared to other memory-based approaches? 
(2) Does ProtoCore achieve higher performance than existing rehearsal-based methods? 
(3) Does ProtoCore yield significant improvements over replay-free CL approaches? 
For coreset-based methods, we include RM~\cite{bang2021rainbow}, PBCS~\cite{zhou2022probabilistic}, BCSR~\cite{hao2023bilevel}, OCS~\cite{yoon2022online}, and CSReL~\cite{tong2025coreset}. 
For replay-based CL methods, we consider iCaRL~\cite{rebuffi2017icarl}, DER~\cite{buzzega2020dark}, DER++~\cite{boschini2022class}, and C-Flat~\cite{bian2024make}. 
For replay-free CL, we include LDC~\cite{gomez2024exemplar}, FeCAM~\cite{goswami2023fecam}, and ADC~\cite{goswami2024resurrecting}.

\begin{figure*}[ht]
    \centering
    \includegraphics[width=\textwidth]{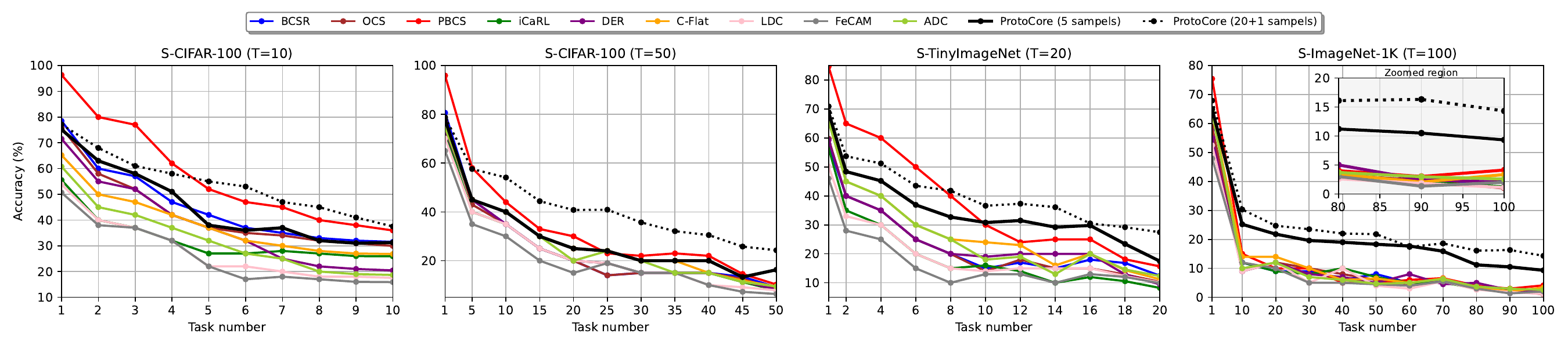}
    \caption{Last accuracy on tasks observed so far in the test set of S-CIFAR-100 (10, 50 tasks), S-TinyImageNet (20 tasks), and S-ImageNet-1K (100 tasks).}
    \label{fig:task-wise-acc}
\end{figure*}
% \begin{figure}[!h]
%      \centering
%      \includegraphics[width=\linewidth]{image-lib/experimental-evaluations/incremental-task/Result1.png}
%      \caption{Last accuracy on tasks observed so far in the test set of ImageNet-1K (10 tasks) and S-TinyImageNet (20 tasks). \TODO{PLACEHOLDER}}
%      \label{fig:task-wise-acc}
% \end{figure}

% \begin{figure}[!h]
%      \centering
%      \includegraphics[width=\linewidth]{image-lib/experimental-evaluations/incremental-task/Result1.png}
%      \caption{Last accuracy on tasks observed so far in the test set of CIFAR-100 (50 tasks) and ImageNet-1K (500 tasks). \TODO{PLACEHOLDER}}
%      \label{fig:task-wise-acc}
% \end{figure}

\noindent\textbf{Dataset:} To evaluate the effectiveness of ProtoCore, particularly under CL settings with long task sequences, we conduct experiments on three standard benchmarks: S-CIFAR-100~\cite{dohare2024loss}, S-TinyImageNet~\cite{hou2019learning}, and S-ImageNet-1K~\cite{dohare2024loss}. We follow the same dataset splits as used in prior works~\cite{hou2019learning, dohare2024loss} to ensure fair comparison. Each benchmark dataset is divided into multiple subsets with non-overlapping label spaces, where each task contains the same number of classes. Specifically, the total number of classes satisfies $C_t \times T = C$, where $C_t$ denotes the number of classes per task, $T$ the number of tasks, and $C$ the total number of classes in the dataset. Please refer to the supplementary material for details on the dataset configurations.

\noindent\textbf{Models:} We use ResNet-18 for S-CIFAR-100, ResNet-50 for S-TinyImageNet, S-ImageNet-1K. Each model is trained with a batch size of 1024 for 200 epochs per task. Please refer to the supplementary material for additional training details on S-CIFAR-100, S-TinyImageNet, and S-ImageNet-1K (see Appendix~\ref{app:settings}). We use VQGAN \cite{} for our pretrained image decoder.

\section{Experimental Evaluation}
% The experiments aim to answer the following questions: 
% \begin{enumerate}
%     \item 
% \end{enumerate}
% \subsection{Motivation Justification}
% \paragraph{Information Loss in Coreset Selection.} \hl{Measure the amount of information loss in coreset/memory selection compared to dataset condensation. The information loss is applied on the same SPC.}

% \paragraph{Forgetting in Generative Replay and Feature Extractor.}

\paragraph{Main Results.}

The result comparisons between ProtoCore and three other categories (i.e., efficient-memory, replay-based, and replay-free CL methods) are presented in Table~\ref{tab:main_results_with_vlines_and_thick_hlines}. In general, ProtoCore demonstrates strong and balanced performance across both short-task and long-task settings.
For short-task scenarios ($T=10,20$), ProtoCore achieves competitive or superior results compared to efficient-memory methods such as BCSR and OnPro, while consistently outperforming replay-based and replay-free baselines. This indicates that ProtoCore effectively retains knowledge and adapts quickly even with limited task sequences.
For long-task scenarios ($T=50,100$), ProtoCore maintains stable performance and shows strong scalability. Despite using only 1–5 stored samples per class, it achieves comparable accuracy to memory-intensive methods like PBCS, and substantially outperforms replay-free approaches (e.g., LDC, FeCAM, ADC). These results highlight ProtoCore's ability to achieve a balance between accuracy and memory efficiency, demonstrating robustness as the number of tasks grows.

\paragraph{Catastrophic forgetting analysis.}
Figure~\ref{fig:task-wise-acc} illustrates how different methods preserve performance on previously learned tasks as new tasks are sequentially introduced. In the short-task settings, i.e., S-CIFAR-100 ($T=10$) and S-TinyImageNet ($T=20$), ProtoCore demonstrates competitive resistance to catastrophic forgetting, closely matching the performance of leading efficient-memory methods such as PBCS and BCSR. In contrast, under the long-task settings, i.e., S-CIFAR-100 ($T=50$) and S-ImageNet-1K ($T=100$), ProtoCore exhibits stronger resistance to forgetting. While baseline methods experience pronounced accuracy degradation after each new task, ProtoCore maintains a notably more stable performance trajectory throughout the continual learning process.

% \subsection{Detailed Analysis}
% \paragraph{Task-wise Feature Extractor Training.} \hl{Plot the T-SNE on test dataset after fixed amount of rounds to show the efficiency and speed of extracting }

% \paragraph{Memory Complexity.} Besides proving the efficient memory over the replay-based methods, we also want to prove our work efficiency over the replay-free methods (e.g., FeCAM, LDC, ADC). We analyze the memory complexity of our proposed approach and compare it with state

% \paragraph{Time complexity.} The primary difference between our method and prior work is that we perform synthetic example generation at the end of every task. Unlike generative memory approaches that optimize full network weights, our method optimizes only the learnable quantized latent vector $z$, which has low dimensionality. Consequently, the optimization required for synthetic generation is substantially faster. To quantify this, we measure the wall-clock time consumed by the synthetic generation procedure per class on CIFAR-100 and report both an empirical per-class time and a simple complexity model. \TODO{Show the time complexity on CIFAR100 to generate per class.}

\subsection{Ablation Tests}
\paragraph{Different choice of loss function.}
\begin{table}[!h]
\caption{Ablation studies of using different losses for $\gL_\text{pro\_cur}$ and $\gL_\text{pro\_pre}$. The contrastive loss is our proposed baselines, while prototypical loss is the loss using by \cite{snell2017prototypical}.
}
\small
\begin{center}
\begin{tabular}{ccccc}
\shline
\multirow{2}{*}{{Method}} & {S-CIFAR-100 (T=10)} & {S-TinyImageNet (T=20)} \\
& Acc $\uparrow$ & Acc $\uparrow$  \\ 
\midrule
Prototypical & 29.51\std{$\pm$0.64} & 15.67\std{$\pm$0.44} \\
Contrastive  & 31.17\std{$\pm$0.83} & 17.31\std{$\pm$0.37} \\
\shline 
\end{tabular}
\end{center}
\label{tab:loss}
\end{table}
Table~\ref{tab:loss} summarizes the ablation results for different prototypical loss functions. As shown, contrastive learning provides a stronger separation between classes, which is reflected in the improved classification accuracy.

\paragraph{Effect of each component.} 
\begin{table}[!h]
\small
\caption{Ablation studies on CIFAR-10 and CIFAR-100. 
ProtoCore denotes the algorithms with all losses. We denote (1) as $\gL_\text{cur\_syn}$, (2) as $\gL_\text{pre\_syn}$, and (3) as $\gL_\text{shift}$ for synthetic exemplar generation. 
We denote (4) as $\gL_\text{pro\_cur}$, (5) as $\gL_\text{pro\_pre}$, and (6) as $\gL_\text{task\_cur}$, (7) as $\gL_\text{task\_pre}$.
}
\begin{center}
\begin{tabular}{ccccc}
\shline
\multirow{2}{*}{Method} & CIFAR-10 & CIFAR-100 \\
& Acc $\uparrow$ (Forget $\downarrow$) & Acc $\uparrow$ (Forget $\downarrow$) \\ 
\midrule
(1) & 43.12\std{$\pm$0.65}(12.5\std{$\pm$0.1}) & 22.43\std{$\pm$0.40}(18.2\std{$\pm$0.5}) \\
(1) + (2)  & 45.45\std{$\pm$0.83}(11.8\std{$\pm$0.3}) & 25.10\std{$\pm$0.18}(16.7\std{$\pm$0.4}) \\
(1) + (2) + (3) & 47.60\std{$\pm$0.10}(10.9\std{$\pm$0.3}) & 27.85\std{$\pm$0.86}(15.5\std{$\pm$0.8}) \\ 
\hline
(4) & 44.20\std{$\pm$0.32}(12.2\std{$\pm$0.6}) & 23.50\std{$\pm$0.39}(17.8\std{$\pm$0.4}) \\
(4) + (5) & 46.10\std{$\pm$0.41}(11.4\std{$\pm$0.3}) & 26.30\std{$\pm$0.77}(16.2\std{$\pm$0.9}) \\ 
\hline
(6) & 42.50\std{$\pm$0.34}(12.8\std{$\pm$0.9}) & 21.90\std{$\pm$0.41}(18.5\std{$\pm$0.5}) \\ 
(6) + (7) & 44.80\std{$\pm$0.63}(12.0\std{$\pm$0.3}) & 24.60\std{$\pm$0.69}(17.0\std{$\pm$0.3}) \\ 
\hline
ProtoCore (ours) & \textbf{49.25}\std{$\pm$0.21}(\textbf{9.8}\std{$\pm$0.3}) & \textbf{37.51}\std{$\pm$0.93}(\textbf{12.0}\std{$\pm$0.7}) \\ 
\shline 
\end{tabular}
\vspace{-1cm}
\end{center}
\label{tab:component}
\end{table}
Table~\ref{tab:component} presents the ablation results for each component of the proposed method.

% \subsubsection{Discarding the false classification samples}

\begin{figure}[!h]
     \centering
     \begin{subfigure}[b]{0.235\textwidth}
         \centering
         \includegraphics[width=\textwidth]{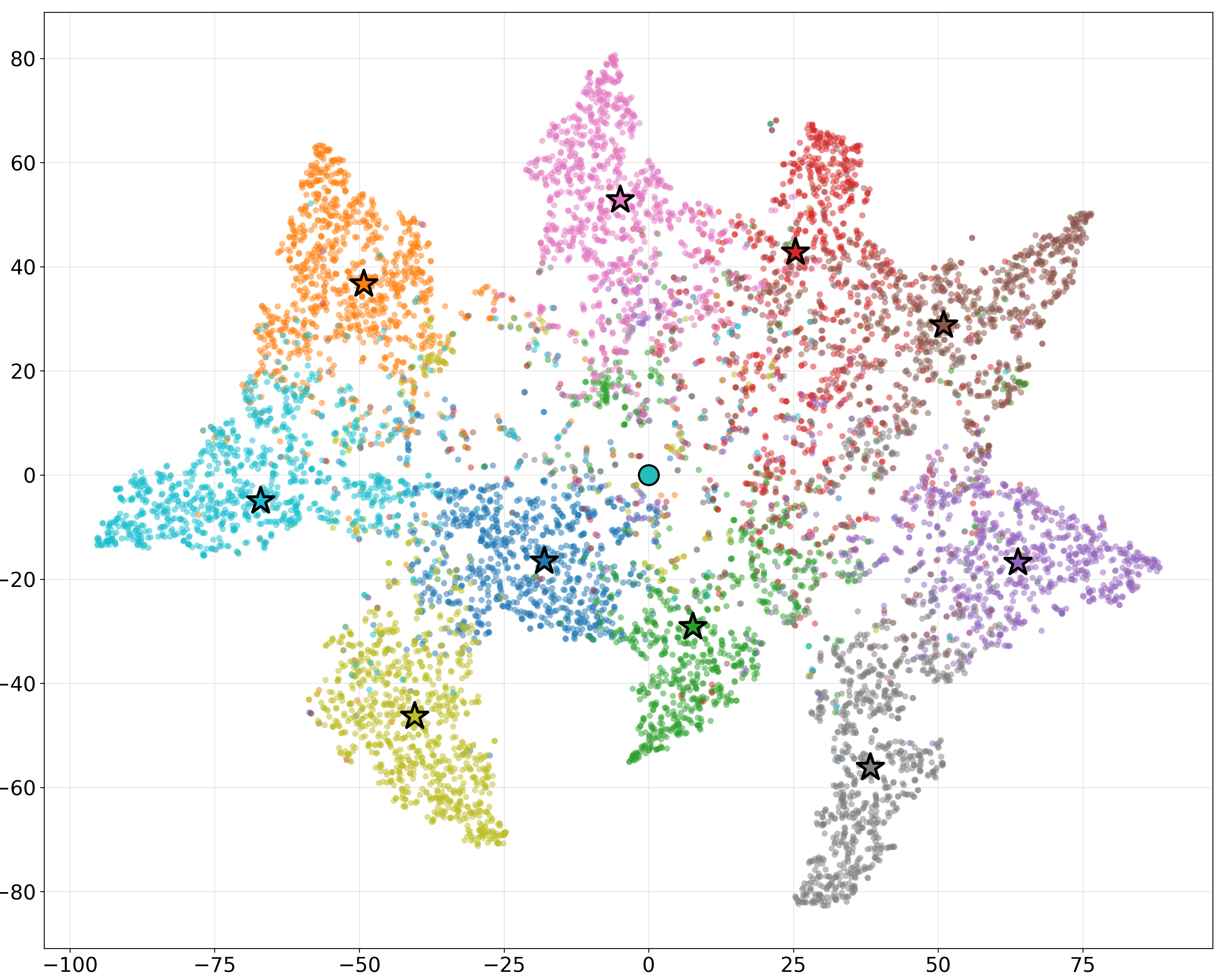}
         \caption{Step 0}
     \end{subfigure}
     \begin{subfigure}[b]{0.235\textwidth}
         \centering
         \includegraphics[width=\textwidth]{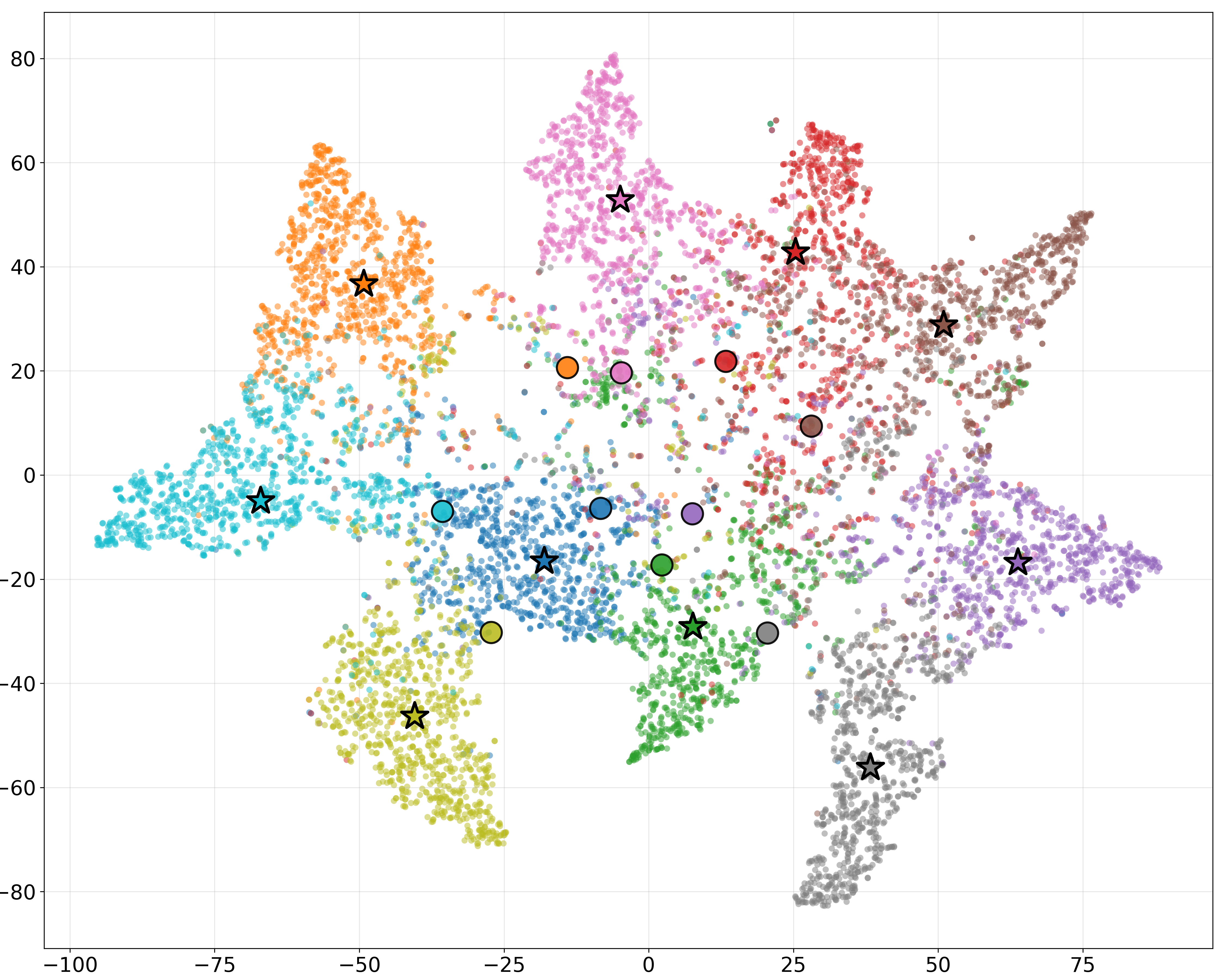}
         \caption{Step 10}
     \end{subfigure}
     \begin{subfigure}[b]{0.235\textwidth}
         \centering
         \includegraphics[width=\textwidth]{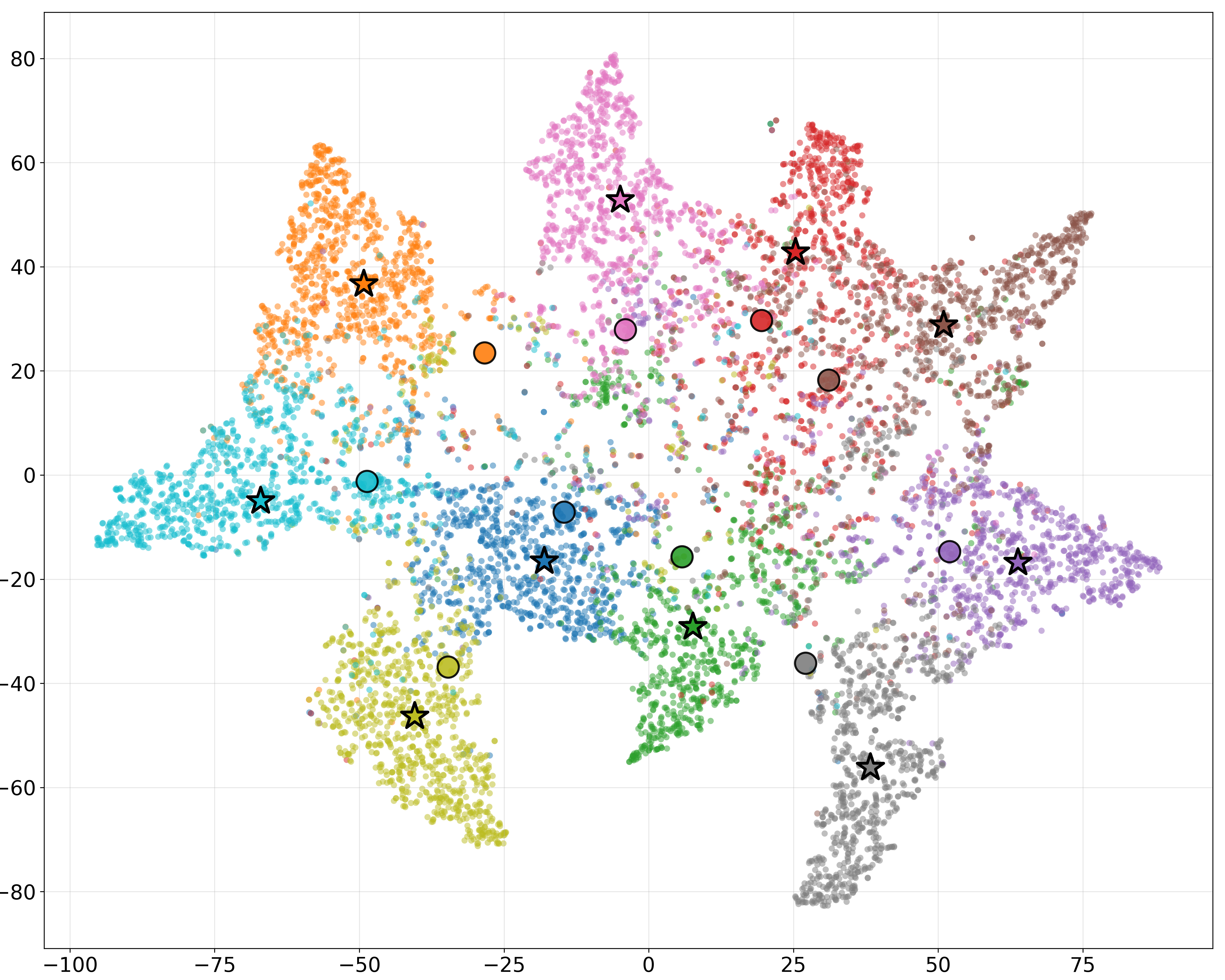}
         \caption{Step 20}
     \end{subfigure}
     \begin{subfigure}[b]{0.235\textwidth}
         \centering
         \includegraphics[width=\textwidth]{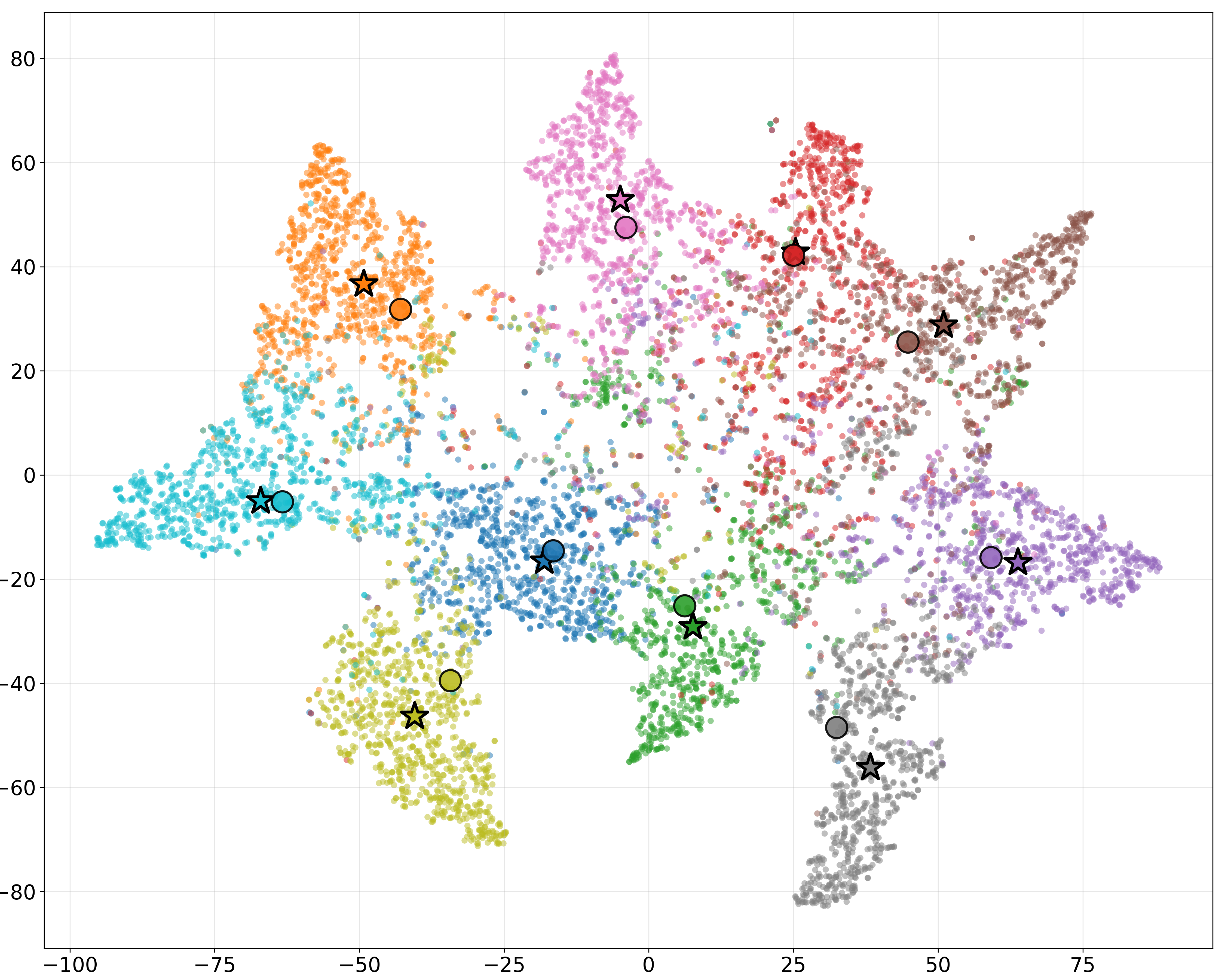}
         \caption{Step 30}
     \end{subfigure}
     \caption{t-SNE visualizations of synthetic features generated by ProtoCore on the CIFAR-100 test set. The synthetic data generation process converges within less than 40 epochs, yielding exemplars that closely approximate the real data distribution. $\bigstar$ represents the real prototypes, $\bullet$ represents the embeddings generated by the synthetic exemplars.}
     \label{fig:synthetic-efficiency}
\end{figure}

\paragraph{Different number of exemplars.}
In Figure \ref{fig:memory_size}, we conducted a comparative analysis on how number of representative exemplars affects the learning accuracy on S-CIFAR-100 under two different task lengths (T=10 and T=50). Overall, performance improves consistently as the number of samples increases, with the most substantial gain occurring when moving from 1 to 5 samples. Notably, using 5 samples achieves the highest accuracy for both settings, while further increasing the memory to 7 samples only brings a marginal improvement. Interestingly, the 1-sample setting, despite being the most memory-efficient, already provides a reasonable accuracy baseline and demonstrates strong efficiency compared to larger memory sizes.
\begin{wraptable}{r}{0.5\linewidth}
    \centering
    \includegraphics[width=\linewidth]{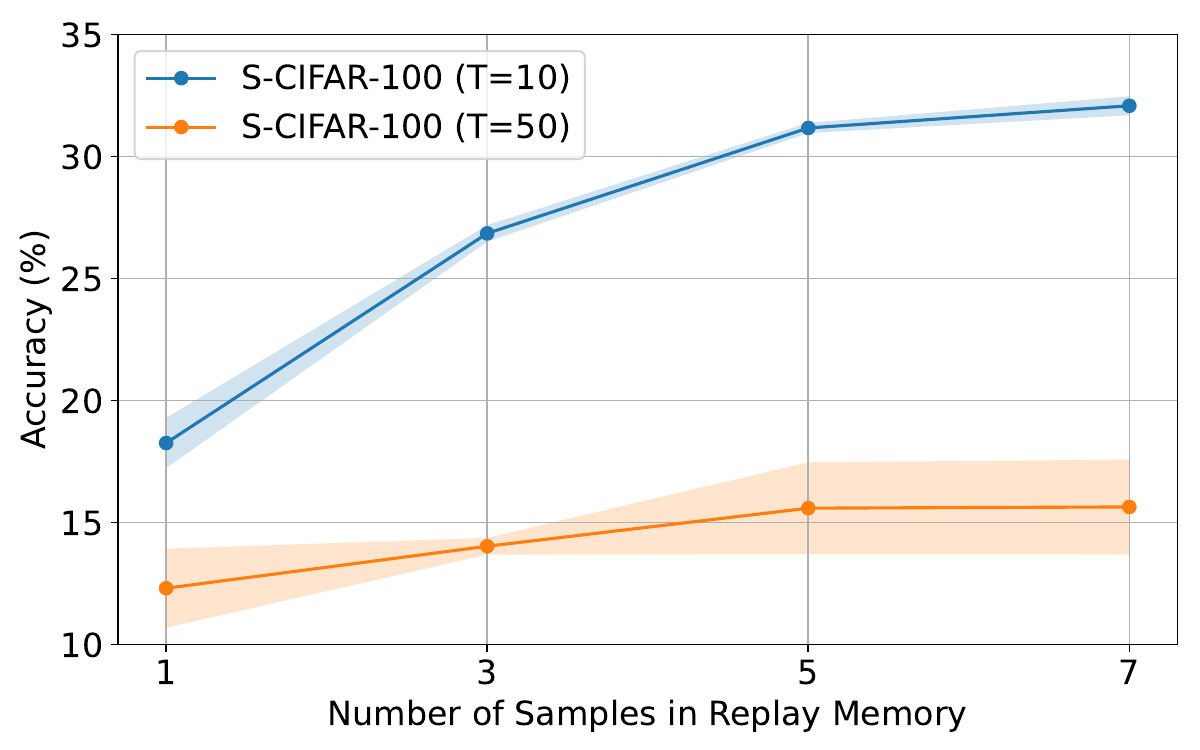}
    \caption{Last accuracy of ProtoCore on S-CIFAR-100 with $T{=}10$ and $T{=}50$ across different replay memory sizes.}
    \label{fig:memory_size}
\end{wraptable}
\paragraph{Task-wise Synthetic Generation.} 
% \hl{Plot the T-SNE to plot the star as exemplars, and triangle as synthetic proto, and X as real proto.}
Figure~\ref{fig:synthetic-efficiency} presents the t-SNE visualization results of synthetic data generation in the first task on the S-CIFAR-100 test set. Several intuitive observations can be made. (1) Certain classes are easily adapted through synthetic prototypes, requiring only a few optimization rounds to align with their real counterparts. (2) Misclassified samples can negatively affect the learning of synthetic prototypes. This occurs because, during batch perturbation, misclassified samples may be aggregated together, forming inaccurate real prototypes. Consequently, the optimization process of the synthetic exemplars may be guided in an incorrect direction.
\paragraph{Analysis of computational and memory costs.}
We conducted a comparative analysis of ProtoCore and baseline methods with respect to computational cost (reported as estimated GFLOPs) and memory usage (GPU and maximum HDD usage), as summarized in Table~\ref{tab:cost_analysis}. 
Regarding memory usage, ProtoCore attains the second-lowest footprint among all baselines. This is expected when compared with replay-based methods, as those approaches store a larger number of samples than ProtoCore ($20$ SPCs compare to $2$ from ProtoCore). Although replay-free methods claim not to retain data from previous tasks, methods such as LDC and ADC store full model checkpoints from earlier tasks. These model snapshots require significantly more memory than exemplar storage, leading to considerably higher overall memory consumption.
The addtional analysis of computational and memory on more challenging data (i.e., S-TinyImageNet) are shown in Appendix~\ref{app:comp-cost}
\begin{table}[!h]\small
\centering
\caption{Comparison on CIFAR-100 in terms of computation (Hours), GPU memory (GB) and memory usage (MB).}
\label{tab:cost_analysis}
\begin{tabular}{lccc}
\toprule
\textbf{Method} & \textbf{Comp. Time}  & \textbf{GPU Mem.}  & \textbf{Max HDD Mem.} \\
                & \textbf{(Hours)}      & \textbf{(GB)}      & \textbf{(MB)}     \\
\midrule
CSReL             & $5.97$ & $1.23$             & $6.20$ \\
BCSR              & $8.16$ & $3.80$             & $6.20$ \\
OCS               & $7.72$ & $3.42$                & $6.20$ \\
PBCS              & $9.79$ & $2.22$             & $6.20$ \\
OnPro             & $8.35$     & $1.75$                & $6.20$ \\
iCaRL             & $0.93$ & $1.31$             & $6.20$ \\
DER               & $1.67$ & $2.10$             & $6.20$ \\
DER++             & $2.10$ & $2.98$             & $6.20$ \\
C-Flat            & $2.51$ & $2.5$                   & $6.20$ \\
LDC               & $1.87$ & $7.06$                   & $46$ \\
FeCAM             & $1.24$ & $2.15$                   & $0$ \\
ADC               & $1.16$ & $6.46$                   & $44$ \\
ProtoCore         & $6.58$ & $4.03$               & $0.26$ \\
\bottomrule
\end{tabular}
\end{table}

\section{Conclusion and Limitations}
\paragraph{Conclusion}
This paper introduces ProtoCore, a framework that synthesizes and stores prototypical exemplars in the experience replay memory as an alternative to conventional data storage. By maintaining only a small number of synthetic samples per class (e.g., 2, 4, or 6), ProtoCore substantially reduces memory requirements while preserving data privacy, since the generated exemplars are not identical to real images. Despite this compression, the framework effectively retains performance in the continual learning setting.
Instead of storing raw prototypes, ProtoCore leverages prototypical exemplars that enable the model to update its feature extractor without relying on original training data. Furthermore, an augmentation mechanism applied to these exemplars enhances their diversity, improving the model’s ability to recall and adapt to previous tasks.
Extensive experiments on widely used benchmark datasets demonstrate the effectiveness of ProtoCore and its individual components. In future work, we plan to explore more efficient strategies, such as incorporating margin-based objectives to further enhance inter-class discriminability.

\paragraph{Limitations} While generating prototypical exemplars demonstrates robustness, a key limitation of both our method and that of \cite{zhu2021prototype} lies in the use of noise for data perturbation. Such noise does not reliably preserve the underlying structure of class clusters, as it often fails to form well-defined, spherical distributions. Mitigating this issue will be an important direction for future work. Additionally, achieving high-quality synthetic exemplars for every task is critical for optimal performance in our approach. Developing strategies to balance exemplar fidelity with computational efficiency remains an open challenge.

\section*{Acknowledgment} 
Minh-Duong Nguyen, Le-Tuan Nguyen, and Dung D. Le are with the Center for AI Research (CAIR), VinUniversity.

%\noindent Kok-Seng Wong is with by the VinUni-Illinois Smart Health Center, VinUniversity.

\noindent Kok-Seng Wong and Dung D. Le are with the Center for Environmental Intelligence (CEI), VinUniversity.
\noindent This work was supported in part by the Green Serverless Computing for Resource-Eﬃcient AI Training Project at VinUniversity under Grant VUNI.CEI.FS 0002;
%\clearpage
{
    \small
    \bibliographystyle{ieeenat_fullname}
    \bibliography{main}
}

\clearpage
\appendix
\clearpage
\setcounter{page}{1}
\maketitlesupplementary

\section{Related Works}\label{sec:related-works}
\subsection{Continual Learning}
Continual Learning (CL) aims to enable a neural network model to learn new tasks continuously without forgetting the old task. Inspired by the working mechanism of the human brain, mainstream memory-based CL methods consolidate previously learned knowledge by replaying old data, thereby avoiding catastrophic forgetting. Due to strict memory and privacy constraints, usually only a small portion of old task data can be kept, many above-mentioned coreset selection methods are used to select informative samples for storage. For example, some heuristic method is used to select the most representative samples from each classes or the sample closest to the decision boundary. 
\paragraph{Experience Replay.}
According to the representative sampling, BCSR \cite{hao2023bilevel} proposes a bilevel optimization framework for coreset selection, where coreset construction is formulated as an upper-level selection problem with model training at the lower level.
OCS \cite{yoon2022online} leverages gradient-based criteria (i.e., minibatch similarity, sample diversity, and coreset affinity) to select a high-quality coreset at each iteration, making the model more robust to noisy and imbalanced data streams.
PBCS \cite{zhou2022probabilistic}
CSReL \cite{tong2025coreset} quantifies the potential performance gain a sample can provide, allowing the method to select high-value samples while avoiding ambiguous or noisy outliers which often degrade performance in traditional bi-level optimization approaches.
According to the boundary-based sampling, RM \cite{bang2021rainbow} propose a diversity-aware sampling method for effectively managing the memory with limited capacity by leveraging classification uncertainty. However, the number of new task's samples and stored old samples in memory is often highly unbalanced, leading the model to be biased towards learning new tasks with more data. 

LPR \cite{yoo2024layerwise} employs a replay-tailored preconditioner applied to the loss gradients. This preconditioner is designed to balance two key objectives: effectively learning from both new and past data, while ensuring that predictions (and internal representations) of past data are modified only gradually, thereby enhancing optimization stability and efficiency.
DEMD \cite{ye2025online} proposes an architecture that integrates a dynamic memory system for retaining the dynamic information alongside a 
PuriDivER \cite{bang2022online} introduces a method for preserving a set of training examples that are both diverse and label-pure. This is achieved through a scoring function that primarily promotes label purity, while incorporating an additional term to encourage diversity by optimizing the sample distribution to resemble that of noisy examples.
RAR \cite{zhang2022repeated} replays stored samples with data augmentation to enhance memory utilization. 

\paragraph{Generative Replay.} Generative replay usually requires training an additional generative model to replay generated data. This is closely related to CL of generative models themselves as they also require incremental updates. 
DGR \cite{shin2017continual} uses a generative model to synthesize realistic past data. Instead of storing actual old samples, the model is jointly trained on new data and the pseudo-samples generated on-demand.
MeRGAN \cite{wu2018memory} employs replay alignment to enforce consistency of the generative data sampled with the same random noise between the old and new generative models, similar to the role of function regularization.

\paragraph{Gradient Replay.} GEM \cite{lopez2017gradient}, LOGD \cite{tang2021layerwise}, MER \cite{riemerlearning} constrain parameter updates to align with the direction of experience replay, corresponding to preserving the previous input space and gradient space with some old training samples.
GPM \cite{saha2021gradient} maintains the gradient subspace important to the old tasks (i.e., the bases of core gradient space) for orthogonal projection in updating parameters. 
VR-MCL \cite{wu2024meta} introduces an approach that implicitly enhances the online Hessian approximation in continual learning by applying a variance reduction technique. This method, which refines the efficiency and stability of weight updates, is designed to mitigate catastrophic forgetting more effectively in non-stationary online data streams.

% \paragraph{Knowledge Distillation based Replay.}

\subsection{Dataset Distillation}
Dataset distillation or condensation aims to condense a large dataset into much smaller yet informative one. It finds applications in neural architecture search, continual learning and privacy protection, etc. 

SRe2L \cite{yin2023squeeze} introduces a three-stage paradigm that leverages highly encoded distribution priors to bypass the need for supervision typically provided during model training. NRR-DD \cite{tran2025enhancing} adopts a distance-based strategy to quantify the discrepancy between predictions on synthetic data and one-hot encoded labels, thereby streamlining the training process and minimizing label storage requirements. Zhu et al. \cite{zhu2023rethinking} observe that dataset distillation often discards semantically meaningful information, and consequently propose masking techniques that enhance calibration with minimal computational overhead. IGD \cite{chen2025influence} is designed to guide diffusion models directly during data distillation under a generalized training-effective condition, eliminating the need for model retraining. Qin et al. \cite{qin2024label} propose a data-knowledge scaling law to quantify the extent to which external knowledge can reduce dataset size. CUDD \cite{ma2025curriculum} introduces a strategic curriculum-based approach for distilling synthetic images, progressing from simple to complex samples. Gu et al. \cite{gu2024efficient} present a dataset distillation framework for diffusion models based on an auxiliary minimax criterion, aimed at enhancing the representativeness and diversity of generated data. Du et al. \cite{du2024diversity} propose a dynamic adjustment mechanism that improves the diversity of synthesized datasets with negligible computational cost, leading to notable performance gains. H-PD \cite{zhong2025hierarchical} systematically explores hierarchical features within the latent space and optimizes them in GANs using loss signals derived from the dataset distillation task. Finally, EDF \cite{wang2025emphasizing} builds on trajectory matching principles by prioritizing updates to discriminative regions, guided by gradient weights extracted from Grad-CAM activation maps.
Deng et al. \cite{deng2024exploiting} introduce the class centralization constraint, which encourages the clustering of samples belonging to the same class in order to enhance class discrimination. To address the limitations of existing methods in aligning feature distributions, they further propose a covariance matching constraint, which facilitates more accurate alignment between real and synthetic datasets by leveraging local feature covariance matrices.

\section{Rethinking of memory in continual learning}\label{sec:motivation}

\begin{comment}
\ks{Approximation errors - When memory is limited and the number of prototypes is fewer than the number of classes or modes, each prototype must represent multiple samples or modes of a class distribution, resulting in coarser approximations of the underlying data. Classes often exhibit multi-modal distributions, e.g., pedestrians in varying poses or lighting conditions and aggressive compression can ignore rare modes, leading to misclassification of unusual inputs. In sublinear or logarithmic growth schemes, prototypes may also be shared across classes or clustered to reduce memory, introducing ambiguity where a single prototype no longer perfectly represents any single class, thereby increasing classification error. Furthermore, in online continual learning, incremental updates with limited memory can amplify prototype drift over time, especially in long task sequences, compounding approximation errors and reducing robustness.}
\end{comment}

\subsection{Scaling of Memory in Continual Learning}
\begin{comment}
\noindent\ks{In real-world continual learning with hundreds/thousands of classes (e.g., ImageNet-scale), how many exemplars per class are feasible?}

\noindent\ks{add a discussion on scaling behavior (linear, sublinear, logarithmic growth of memory) to strengthen practical motivation - practical memory savings.}

\begin{itemize}
\color{red}
    \item Different growth policies affect accuracy, forgetting, and computational cost. E.g., Linear growth gives high accuracy but is memory-heavy; sublinear/logarithmic growth is efficient but risks approximation errors.
    \begin{enumerate}
        \item linear - memory grows proportionally to the number of classes $C$. 
        \item sublinear - memory grows slower than the number of classes, e.g., $O\sqrt{C}$.
        \item logarithmic - memory grows very slowly, e.g., $O(log(C))$.
    \end{enumerate}
    \item Explain how exemplar condensation can reduce memory from linear to sublinear or logarithmic growth.
\end{itemize}
\end{comment}

A central challenge in CL lies in the scalability of memory usage as the number of tasks grows. 
One of the standard approaches for managing the replay buffer is reservoir sampling \cite{buzzega2020dark, boschini2022class}, where incoming data are inserted into the buffer with a uniform probability, and existing exemplars are randomly discarded once the memory is full. However, this method has been shown to suffer from a noticeable decline in performance, particularly when the number of tasks is relatively small (fewer than 20) \cite{boschini2022class}. This degradation mainly arises from the random nature of exemplar replacement, which can lead to the premature removal of informative samples and reduce the representational quality of the stored buffer.

Alternatively, coreset selection strategies allocate a fixed number of exemplars per class \cite{bang2021rainbow, tong2025coreset}. 
This approach generally achieves significantly better performance than reservoir sampling, since the memory is constructed from samples that are more salient to the learning process. 
However, these methods typically incur linear growth in memory requirements with respect to the number of tasks, i.e., $\mathcal{O}(C N_C)$, where $C$ denotes the number of classes and $N_C$ the number of stored samples per class. 
In practice, coreset selection usually requires up to $100$ exemplars per class to maintain competitive performance \cite{bang2021rainbow}. 
When the number of exemplars per class is reduced to around $20$, however, the model suffers a substantial accuracy drop by $20\%$ according to empirical observations \cite{borsos2020coresets}.
This strategy, while effective in mitigating catastrophic forgetting, becomes impractical in real-world deployments where tasks and number of classes $C$ accumulate over time and storage is constrained, as is common in edge devices and mobile platforms.

To address this limitation, we shift our focus toward prototype-based memory, which aims to minimize the number of stored samples per class while preserving the representational capacity needed for effective continual learning.

\subsection{Prototypes in Continual Learning}
Fundamentally, prototype-based memory \cite{chen2023saving,wei2023online,asadi2023prototype} has gained the big interest to further enhance memory efficiency thanks to several advantages: 
\begin{enumerate}
    \item prototypes offer a lightweight alternative. Instead of storing entire data samples, we can reduce the memory footprint significantly \cite{chen2023saving}.
    \item Since the prototype captures the central tendency of the data within each class distribution, it can naturally facilitate data augmentation through simple transformation techniques, such as Gaussian noise injection \cite{zhu2021prototype}. Consequently, synthetic subsets can be generated by perturbing samples around the prototype, effectively producing diverse data points from a single exemplar.
    \item Prototypes provide a stable reference point for previously learned knowledge. By maintaining these prototypes, the model can preserve the structure of the embedding space. When learning a new task, the model is encouraged to keep new class embeddings close to their own prototypes while maintaining a clear distance from the prototypes of old classes. This approach not only minimizes storage requirements but also maintains high performance in continual learning scenarios. 
\end{enumerate}

Although prototypes have demonstrated robustness in CL, their use remains underexplored in current research, primarily due to several limiting factors:
\begin{enumerate}
    \item \textbf{Prototypes as guidance for rehearsal replay.} When prototypes are employed in rehearsal-based replay, they primarily serve two purposes. The first is to guide the optimization of the feature extractor \cite{ho2023prototype, lin2022prototype}. However, this approach does not fully replace traditional rehearsal methods that store raw samples, as the guidance process still relies on previous samples to compute the necessary updates.
    
    \item \textbf{Prototypes as stored exemplars.} When prototypes are directly used as stored exemplars for replay, the backpropagation process is applied only to the classifier layer \cite{chen2023saving, jiangpeng2022exemplar}. Consequently, the feature extractor cannot be effectively updated, which limits the model's capacity to adapt and retain knowledge across tasks.
    
    \item \textbf{Vulnerability to catastrophic forgetting.} The construction of prototypes depends on effectively trained prototypical networks. However, similar to other parametric models, prototypical networks are susceptible to catastrophic forgetting \cite{lange2021continual, wei2023online}. As new tasks are introduced, previously learned prototypes may be overwritten or drift within the representation space, leading to a substantial performance decline on earlier tasks. This problem becomes increasingly severe as the number of classes grows or the task sequence lengthens.
\end{enumerate}
% \subsection{Dataset Distillation for Prototype-Based Memory}

\section{Notations}
\begin{table}[!h]
\centering
\caption{Notations used in the proposed algorithms.}
\begin{tabular}{lp{0.65\linewidth}}
\hline
\textbf{Notation} & \textbf{Description} \\
\hline
$T$ & Total number of tasks. \\
$\vert\gT_t\vert$ & Number of samples per task $t$. \\
$\gT_t$ & Dataset of task $t$ with $\vert\gT_t\vert$ samples. \\
$\gC_t$ & Set of available class for task $t$. \\
$\gC$ & Set of available class. \\
$y_i^t$ & Class label of sample $x_i^t$. \\
$f_\theta(\cdot)$ & Feature extractor with parameters $\theta$. \\
$f_\phi(\cdot)$ & Classifier with parameters $\phi$. \\
$g(\cdot)$ & Pretrained image decoder. \\
$\gM$ & Memory buffer storing real exemplars $\gM_x$ and synthetic exemplars $\gM_s$. \\
$\gM_x, \gM^c_x$ & Memory of all real exemplars, and real exemplars according to class $c$. \\
$\gM_s$ & Memory of synthetic exemplars. \\
$\gM_p$ & Memory of real prototypes. \\
$\gB_m$ & Batch sampled from memory at previous task $m$. \\
$s^c$ & Currently trained synthetic exemplar for class $c$. \\
$\hat{s}^c$ & Stored synthetic exemplar for class $c$. \\
$p^c$ & Prototype of class $c$ computed from real exemplars of the current class $t$. \\
$\hat{p}^c$ & Prototype of class $c$ computed from synthetic exemplars in $M_s$. \\
% $\gL_{\text{CE}}$ & Cross-entropy loss on current task data \\
% $\gL_{\text{proto}}$ & Prototypical loss (aligning features with class prototypes) \\
% $\gL_{\text{proto\_align}}$ & Alignment loss between real and synthetic prototypes \\
% $\gL_{\text{\mathtt{sync\_proto}}}$ & Synchronization loss for synthetic exemplar optimization \\
% $\gL_{\text{\mathtt{total}}}$ & Total training loss) \\
$\eta$ & Learning rate. \\
$E$ & Number of training epochs per task. \\
\hline
\end{tabular}
\end{table}

\clearpage
\section{Algorithm Details}
We present the detailed procedure of ProtoCore in Algorithm~\ref{alg:ProtoCore-only}. We present the detailed procedure of ProtoCore + replayed-based CL in Algorithm~\ref{alg:ProtoCore-memory}
\subsection{Full Memory Replay}

\begin{algorithm}[!h]
    \DontPrintSemicolon
    \SetKwInput{KwInput}{Input}
    \SetKwInput{KwOutput}{Output}
    % \SetKwProg{compensate}{Function \emph{compensate}}{}{end}
    \KwInput{Feature extractor \(f_{\theta}\), classifier \( f_{\phi}\), pretrained image encoder $g$, number of tasks \( T \), training set \( \gT = \{\{(x_i^t, y_i^t)\}_{i=1}^{\vert\gT_t\vert}\}_{t=1}^T \), \( y_i^t \in \mathcal{Y} \), $\gM = \gM_s \bigcup \gM_x \bigcup \gM_p$.}
    \KwOutput{$\theta^{T}$;}
    % \tcp*[l]{initialization}\
    % \( f_{\theta} \), \( f_{\phi} \), \( \gM = \emptyset \) \\
    \tcp*[l]{iterative approximation} \
    \For{$t$ in $1,...,T$}{
        Obtain past model $\theta' \gets \theta^{t-1}$. \\
        Obtain current task data $\gT_t=\{(x_i^t, y_i^t)\}_{i=1}^{\vert\gT_t\vert}$. \\
        Obtain previous task data $\{\{(\hat{x}_i^t, \hat{y}_i^t)\}_{i=1}^{\vert\gB_m\vert}\}_{m=1}^{t-1} \gets \gM_x$. \\
        % Combine current task and memory pool \( \bar{\gT} = \gT_t \cup \gM \). \\
        \tcp*[l]{Init learnable data at begin of every task} \
        Initialize learnable synthetic latent $z$, optimizer $\mathtt{opt}(z)$, image decoder $s = g(z)$. \\
        \For{$e$ in $0,\ldots,E-1$}{
            \For{ batch in \(\gT_t\)}{
                Obtain $(x, y)$ from batch. \\ 
                {\color{blue}\tcp*[l]{Loss Computation}}\
                Compute $\gL_{\mathtt{total}}$ according to Alg.~\ref{alg:continual-exemplar-alignment}. \\
                {\color{blue}\tcp*[l]{Model update}}\
                $\theta{'} = \theta{'} - \eta \nabla_{\theta'} \gL_{\mathtt{total}}$\\
                % \( \gM_x \gets reservoir(\gM, (x, y)) \). \\
                % \uIf{$e = E-1$}{
                    % }
            }
        }
        $\theta^{t} \gets \theta'$. \\
        \For{$e$ in $0,\ldots,E-1$}{
            \For{ sample in \(\gT_t\)}{
                Obtain $(x, y)$ from sample. \\ 
                {\color{orange}\tcp*[l]{Prototypical Exemplar Loss Computation}}\
                Compute $\gL_{\mathtt{sync\_proto}}$ according to Alg.~\ref{alg:proto-exemplar}. \\
                {\color{orange}\tcp*[l]{Optimize Prototypical Exemplar}}\
                $s = s - \eta \nabla_s \gL_{\mathtt{sync\_proto}}$\\
                % \( \gM_x \gets reservoir(\gM, (x, y)) \). \\
                % \uIf{$e = E-1$}{
                    % }
            }
        }        
        \tcp*[l]{Store exemplars and data.}\
        \( \gM_x \gets \mathtt{reservoir}(\gM, (x, y)\in\gT_t) \). \\
        \( \gM_s \gets \mathtt{replace}(\gM, (s, y)) \). \\
        \( \gM_p \gets \mathtt{replace}(\gM, (p)) \). \\
    }
    \caption{ProtoCore (Replay memory supported). The {\color{blue}blue} annotations reveals the model update of continual learning. The {\color{orange}orange} annotations demonstrates the prototypical exemplar generation.}
    \label{alg:ProtoCore-memory}    
\end{algorithm}

\begin{algorithm}[htbp]
    \DontPrintSemicolon
    \SetKwInput{KwInput}{Input}
    \SetKwInput{KwOutput}{Output}
    % \SetKwProg{compensate}{Function \emph{compensate}}{}{end}
    \KwInput{Feature extractor \(f_{\theta}\), classifier \( f_{\phi}\), current task \( \gT_t \), \( y_i^t \in \mathcal{Y} \), $\gM = \gM_s \bigcup \gM_x \bigcup \gM_p$. Coefficients $\alpha_1, \alpha_2, \alpha_3, \beta_1, \beta_2$.}
    \KwOutput{$\gL_{\mathtt{total}}$;}
    \tcp*[l]{initialization}\
    \For{$c$ in $\gC^t$ seen classes}{
        Samples $\{\hat{x}^c_i\}^{\vert\gM^{c}_x\vert}_{i=1}$ according to class $c$ in $\gM_x$.\\
        Samples $\hat{s}^c$ according to classes $c$ in $\gM_s$.\\
        % Samples $\hat{p}^c$ according to classes $c$ in $\gM_p$.\\
    }  
    {\color{blue}\tcp*[l]{Task-head Learning with Memory}} \
    {\color{blue}\tcp*[l]{Current Task}} \
    $\gL_{\mathtt{task\_cur}} = 0$. \\
    \For{$(x;y)$ in $\gT_t$}{
        % $z=f_{\theta}(x)$. \\ 
       \(\gL_{\mathtt{task\_cur}}~\mathrel{+}= \mathtt{CE}(f_{\phi}(f_{\theta}(x)), y) \). \\
    }
    {\color{blue}\tcp*[l]{Previous Tasks}} \
    % $\gL_{\mathtt{task\_pre}} = 0$. \\
    $\mathbf{z} = \{f_{\theta}(h(s)) \vert \forall h\in\gF\}$. \\ 
    \(\gL_{\mathtt{task\_pre}}~\mathrel{+}= \mathtt{CE}(f_{\phi}(\mathbf{z}), y) \). \\
    {\color{blue}\tcp*[l]{Full Memory Replay}} \
    \For{$(x;y)$ in $\gM_x$}{
        % $z=f_{\theta}(x)$. \\ 
       \(\gL_{\mathtt{task\_pre}}~\mathrel{+}= \mathtt{CE}(f_{\phi}(f_{\theta}(x)), y) \). \\
    }
    {\color{orange}\tcp*[l]{Incremental Feature Extractor}} \
    {\color{orange}\tcp*[l]{Assign prototypes to list}}\ 
    \For{$c$ in $\gC^{1:t-1}$ seen classes}{
        Samples $\hat{s}^c$ according to classes $c$ in $\gM_s$.\\
        \uIf{$c$ is in $\gC^t$ current classes}{
            Samples $\gT^c_t$ according to classes $c$ in $\gT_t$.\\
        $p^{c} 
        = \frac{\beta_1}{\vert \gT^c_t\vert}\sum_{(x;\cdot)\sim\gT^c_t}f_{\theta}(x)
        + \frac{\beta_2}{\vert\gF\vert}\sum_{h\sim\gF}f_{\theta}(h(\hat{s}^c))$. \\ 
        }
        \Else{
            $p^{c} = \frac{\beta_2}{\vert\gF\vert}\sum_{h\sim\gF}f_{\theta}(h(\hat{s}^c))$. \\ 
        }
    }
    % {\color{orange}\tcp*[l]{Current Task}} \
    % $\gL_{\mathtt{cur\_pro}} = 0$. \\
    % \For{$c$ in $\gC^{t}$ current classes}{
    %     \For{$(x;y)$ in $\gT_{1:t-1}$ according to $c$}{
    %         \( \gL_{\mathtt{cur\_pro}} \mathrel{+}= \frac{\exp{(-\mathtt{MSE}(f_{\theta}(x);p^c))}}{\sum_{l{'}\neq l}\exp{(-\mathtt{MSE}(f_{\theta}(x);p^{l{'}}))}} \).
    %     }
    % }
    % {\color{orange}\tcp*[l]{Exemplar Alignment Loss}} \
    % \(\gL_{\mathtt{pre\_pro}} = 0 \). \\
    % \For{$c$ in $\gC^{1:t-1}$ seen classes}{
    %     % Samples $\{(\hat{x}^c_i, {\hat{y}^c_i})\}^{n^{c}}_{i=1}$ according to classes $c$ in $\gM_x$.\\
    %     Samples $\gM^c_x$ according to classes $c$ in $\gM_x$.\\
    %     Samples $\gT^c_t$ according to classes $c$ in $\gT_t$.\\
    %     \(\gL_{\mathtt{pre\_pro}} \mathrel{+}= \mathtt{MSE}(\hat{p}^{c}; f_{\theta}(\hat{s}^c)) \).
    % }
    {\color{orange}\tcp*[l]{Prototypical Loss}} \
    $\gL_{\mathtt{cur\_pro}}, \gL_{\mathtt{pre\_pro}} = 0,0$. \\
    % $\gL_{\mathtt{pre\_pro}} = 0$. \\
    \For{$c$ in $\gC^t$ seen classes}{
        \uIf{$c$ is in $\gC^t$ current classes}{
            \For{$(x;y)$ in $\gT_t$ according to $c$}{
                \( \gL_{\mathtt{cur\_pro}} \mathrel{+}= \frac{\exp{(-\mathtt{MSE}(f_{\theta}(x);p^c))}}{\sum_{l{'}\neq l}\exp{(-\mathtt{MSE}(f_{\theta}(x);p^{l{'}}))}} \).
            }
        }
        \Else{
            \For{$(s;y)$ in $\gM_s$}{
                $\mathbf{z} = \{f_{\theta}(h(s)) \vert \forall h\in\gF\}$
                \( \gL_{\mathtt{pre\_pro}} \mathrel{+}= \frac{\exp{(-\mathtt{MSE}(\mathbf{z};p^c))}}{\sum_{l{'}\neq l}\exp{(-\mathtt{MSE}(\mathbf{z};p^{l{'}}))}} \). \\
            }
            \For{$(x;y)$ in $\gM_x$ according to $c$}{
                \( \gL_{\mathtt{pre\_pro}} \mathrel{+}= \frac{\exp{(-\mathtt{MSE}(f_{\theta}(x);p^c))}}{\sum_{l{'}\neq l}\exp{(-\mathtt{MSE}(f_{\theta}(x);p^{l{'}}))}} \).
            }
        }
    }
    \Return{$\gL_{\mathtt{total}} = \gL_{\mathtt{cur\_pro}} + \alpha_1\gL_{\mathtt{pre\_pro}} + \alpha_2\gL_{\mathtt{task\_pre}} + \alpha_3\gL_{\mathtt{task\_cur}}$}\\
    \caption{Network update with prototypical exemplar alignment (Replay memory supported). The {\color{blue}blue} annotations demonstrates the task-head learning. The {\color{orange}orange} annotations refers to the prototype learning for the feature extractor.}
    \label{alg:continual-exemplar-alignment}    
\end{algorithm}

\begin{algorithm}[htbp]
    \DontPrintSemicolon
    \SetKwInput{KwInput}{Input}
    \SetKwInput{KwOutput}{Output}
    % \SetKwProg{compensate}{Function \emph{compensate}}{}{end}
    \KwInput{Feature extractor \(f_{\theta}\), classifier \( f_{\phi}\), current task \( \gT_t \), \( y_i^t \in \mathcal{Y} \), synthetic data $s = \{s^c\}^{\gC^{t}}_{l=1}$ and optimizer $\mathtt{opt}(s)$, coefficient $\alpha_1, \alpha_2$, $\gM = \gM_s \bigcup \gM_x \bigcup \gM_p$.}
    \KwOutput{$\theta^{t+T}$;}
    \tcp*[l]{Current Exemplar Alignment}\
    \For{$c$ in current classes of $\gT_t$}{
        Samples $\gT^c_t$ according to classes $c$ in $\gT_t$.\\
        $p^{c} = \frac{1}{n^{c}}\sum_{(x,y)\sim\gT^c_t}f_{\theta}(x)$. \\ 
        \(\gL_{\mathtt{cur\_syn}} \mathrel{+}= \mathtt{MSE}(f_{\theta}(s^c); p^{c}) \). \\
    }  
    \tcp*[l]{Previous Exemplar Alignment} \
    \For{$c$ in $\gC^{1:t-1}$ seen classes}{
        \uIf{$c$ is in current classes}{
            Samples $\hat{s}^c$ according to classes $c$ in $\gM_s$.\\
            \(\gL_{\mathtt{pre\_syn}} \mathrel{+}= \mathtt{MSE}(f_{\theta}(s^c); f_{\theta}(\hat{s}^c)) \). \\ 
            % $\gL_{\mathtt{sync\_proto}} \mathrel{+}= \frac{\exp{(-\mathtt{MSE}(f_{\theta}(s);p^c))}}{\sum_{l{'}\neq l}\exp{(-\mathtt{MSE}(f_{\theta}(s);p^{l{'}}))}}$. \\    
        }
    }
    \tcp*[l]{Representation Shift Alignment} \
    \For{$c$ in $\gC^{1:t-1}$ seen classes}{
        \uIf{$c$ is in current classes}{
            Samples $\hat{s}^c$ according to classes $c$ in $\gM_s$.\\
            Samples $\hat{p}^c$ according to classes $c$ in $\gM_p$.\\
            \(\gL_{\mathtt{shift}} \mathrel{+}= \mathtt{MSE}(f_{\theta}(s^c); \hat{p}^c) \). \\ 
        }
    }
    \Return{$\gL_{\mathtt{sync\_proto}} = \gL_{\mathtt{cur\_syn}} + \alpha_1\gL_{\mathtt{pre\_syn}} + \alpha_2\gL_{\mathtt{shift}}$}\\
    \caption{Prototypical Exemplar Generation.}
    \label{alg:proto-exemplar}    
\end{algorithm}

\clearpage
\subsection{Prototypical-Exemplar Only Replay}

\begin{algorithm}[!h]
    \DontPrintSemicolon
    \SetKwInput{KwInput}{Input}
    \SetKwInput{KwOutput}{Output}
    % \SetKwProg{compensate}{Function \emph{compensate}}{}{end}
    \KwInput{Feature extractor \(f_{\theta}\), classifier \( f_{\phi}\), number of tasks \( T \), data \( \gT = \{\{(x_i^t, y_i^t)\}_{i=1}^{N_t}\}_{t=1}^T \), \( y_i^t \in \mathcal{Y} \), memory pool $\gM = \gM_s$.}
    \KwOutput{$\theta^{T}$;}
    \tcp*[l]{initialization}\
    \( f_{\theta} \), \( f_{\phi} \), \( \gM = \emptyset \) \\
    \tcp*[l]{iterative approximation} \
    \For{$t$ in $1,...,T$}{
        Obtain past model $\theta' \gets \theta^{t-1}$. \\
        Obtain current task data $\gT_t=\{(x_i^t, y_i^t)\}_{i=1}^{n_t}$. \\
        % Combine current task and memory pool \( \bar{\gT} = \gT_t \cup \gM \). \\
        \tcp*[l]{Init learnable data at begin of every task} \
        Initialize learnable synthetic data $s$, optimizer $\mathtt{opt}(s)$. \\
        \For{$e$ in $0,\ldots,E-1$}{
            \For{ sample in \(\gT_t\)}{
                Obtain $(x, y)$ from sample. \\ 
                {\color{blue}\tcp*[l]{Loss Computation}}\
                Compute $\gL_{\mathtt{total}}$ according to Alg.~\ref{alg:continual-exemplar-alignment}. \\
                {\color{blue}\tcp*[l]{Model update}}\
                $\theta{'} = \theta{'} - \eta \nabla_{\theta'} \gL_{\mathtt{total}}$\\
                % \( \gM_x \gets reservoir(\gM, (x, y)) \). \\
                % \uIf{$e = E-1$}{
                    % }
            }
        }
        $\theta^{t} \gets \theta'$. \\
        \For{$e$ in $0,\ldots,E-1$}{
            \For{ sample in \(\gT_t\)}{
                Obtain $(x, y)$ from sample. \\ 
                {\color{orange}\tcp*[l]{Prototypical Exemplar Loss Computation}}\
                Compute $\gL_{\mathtt{sync\_proto}}$ according to Alg.~\ref{alg:proto-exemplar}. \\
                {\color{orange}\tcp*[l]{Optimize Prototypical Exemplar}}\
                $s = s - \eta \nabla_s \gL_{\mathtt{sync\_proto}}$\\
                % \( \gM_x \gets reservoir(\gM, (x, y)) \). \\
                % \uIf{$e = E-1$}{
                    % }
            }
        }        
        \tcp*[l]{Store exemplars and data.}\
        \( \gM_s \gets replace(\gM, (s, y)) \). \\
    }
    \caption{ProtoCore (Synthetic Replay Only). The {\color{blue}blue} annotations reveals the model update of continual learning. The {\color{orange}orange} annotations demonstrates the prototypical exemplar generation.}
    \label{alg:ProtoCore-only}    
\end{algorithm}

\begin{algorithm}[!h]
    \DontPrintSemicolon
    \SetKwInput{KwInput}{Input}
    \SetKwInput{KwOutput}{Output}
    % \SetKwProg{compensate}{Function \emph{compensate}}{}{end}
    \KwInput{Feature extractor \(f_{\theta}\), classifier \( f_{\phi}\), current task \( \gT_t \), \( y_i^t \in \mathcal{Y} \), memory $\gM = \gM_s$. Coefficients $\alpha_1, \alpha_2, \alpha_3, \beta$, Transformation set $\gF$. }
    \KwOutput{$\gL_{\mathtt{total}}$;}
    \tcp*[l]{initialization}\
    \For{$c$ in $\gC^t$ seen classes}{
        Samples $\hat{s}^c$ according to classes $c$ in $\gM_s$.\\
    }  
    {\color{blue}\tcp*[l]{Task-head Learning with Memory}} \
    {\color{blue}\tcp*[l]{Current Task}} \
    $\gL_{\mathtt{task\_cur}} = 0$. \\
    \For{$(x;y)$ in $\gT_t$}{
        $z=f_{\theta}(x)$. \\ 
       \(\gL_{\mathtt{task\_cur}}~\mathrel{+}= \gL_{CE}(f_{\phi}(z), y) \). \\
    }
    {\color{blue}\tcp*[l]{Previous Tasks}} \
    $\gL_{\mathtt{task\_pre}} = 0$. \\
    $\mathbf{z} = \{f_{\theta}(h(s)) \vert \forall h\in\gF\}$. \\ 
    \(\gL_{\mathtt{task\_pre}}~\mathrel{+}= \mathtt{CE}(f_{\phi}(\mathbf{z}), y) \). \\
    {\color{orange}\tcp*[l]{Incremental Feature Extractor}}\ 
    {\color{orange}\tcp*[l]{Assign prototypes to list}}\ 
    \For{$c$ in $\gC^{1:t-1}$ seen classes}{
        Samples $\hat{s}^c$ according to classes $c$ in $\gM_s$.\\
        \uIf{$c$ is in $\gC^t$ current classes}{
            Samples $\gT^c_t$ according to classes $c$ in $\gT_t$.\\
            $p^{c} = \frac{\beta}{\vert\gF\vert}\sum_{h\sim\gF}f_{\theta}(h(\hat{s}^c))
            +\frac{1-\beta}{\vert \gT^c_t\vert}\sum_{(x;\cdot)\sim\gT^c_t}f_{\theta}(x)$. \\ 
        }
        \Else{
            $p^{c} = \frac{1}{\vert\gF\vert}\sum_{h\sim\gF}f_{\theta}(h(\hat{s}^c))$. \\ 
        }
    }
    {\color{orange}\tcp*[l]{Prototypical Loss}} \
    $\gL_{\mathtt{cur\_pro}} = 0$. \\
    $\gL_{\mathtt{pre\_pro}} = 0$. \\
    \For{$c$ in $\gC^t$ seen classes}{
        \uIf{$c$ is in $\gC^t$ current classes}{
            \For{$(x;y)$ in $\gT_t$ according to $c$}{
                \( \gL_{\mathtt{cur\_pro}} \mathrel{+}= \frac{\exp{(-\mathtt{MSE}(f_{\theta}(x);p^c))}}{\sum_{l{'}\neq l}\exp{(-\mathtt{MSE}(f_{\theta}(x);p^{l{'}}))}} \).
            }
        }
        \Else{
            \For{$(s;y)$ in $\gM_s$}{
                $\mathbf{z} = \{f_{\theta}(h(s)) \vert \forall h\in\gF\}$
                \( \gL_{\mathtt{pre\_pro}} \mathrel{+}= \frac{\exp{(-\mathtt{MSE}(\mathbf{z};p^c))}}{\sum_{l{'}\neq l}\exp{(-\mathtt{MSE}(\mathbf{z};p^{l{'}}))}} \). \\
            }
        }
    }
    \Return{$\gL_{\mathtt{total}} = \gL_{\mathtt{cur\_pro}} + \alpha_1\gL_{\mathtt{pre\_pro}} + \alpha_2\gL_{\mathtt{task\_pre}} + \alpha_3\gL_{\mathtt{task\_cur}}$}\\
    \caption{Network update with prototypical exemplar alignment (Synthetic Replay Only). The {\color{blue}blue} annotations demonstrates the task-head learning. The {\color{orange}orange} annotations refers to the prototype learning for the feature extractor.}
    \label{alg:continual-exemplar-alignment-2}    
\end{algorithm}

\begin{algorithm}[!h]
    \DontPrintSemicolon
    \SetKwInput{KwInput}{Input}
    \SetKwInput{KwOutput}{Output}
    % \SetKwProg{compensate}{Function \emph{compensate}}{}{end}
    \KwInput{Feature extractor \(f_{\theta}\), classifier \( f_{\phi}\), current task \( \gT_t \), \( y_i^t \in \mathcal{Y} \), synthetic data $s = \{s^c\}^{\gC^{t}}_{l=1}$ and optimizer $\mathtt{opt}(s)$, coefficient $\alpha$.}
    \KwOutput{$\theta^{t+T}$;}
    \tcp*[l]{Current Exemplar Alignment}\
    \For{$c$ in current classes of $\gT_t$}{
        Samples $\gT^c_t$ according to classes $c$ in $\gT_t$.\\
        $p^{c} = \frac{1}{n^{c}}\sum_{(x,y)\sim\gT^c_t}f_{\theta}(x)$. \\ 
        \(\gL_{1} \mathrel{+}= \mathtt{MSE}(f_{\theta}(s^c); p^{c}) \). \\
    }  
    \tcp*[l]{Previous Exemplar Alignment} \
    \For{$c$ in $\gC^{1:t-1}$ seen classes}{
        \uIf{$c$ is in current classes}{
            Samples $\hat{s}^c$ according to classes $c$ in $\gM_s$.\\
            \(\gL_{2} \mathrel{+}= \mathtt{MSE}(f_{\theta}(s^c); f_{\theta}(\hat{s}^c)) \). \\ 
            % $\gL_{\mathtt{sync\_proto}} \mathrel{+}= \frac{\exp{(-\mathtt{MSE}(f_{\theta}(s);p^c))}}{\sum_{l{'}\neq l}\exp{(-\mathtt{MSE}(f_{\theta}(s);p^{l{'}}))}}$. \  
        }
    }
    \tcp*[l]{Representation Shift Alignment} \
    \For{$c$ in $\gC^{1:t-1}$ seen classes}{
        \uIf{$c$ is in current classes}{
            Samples $\hat{s}^c$ according to classes $c$ in $\gM_s$.\\
            Samples $\hat{p}^c$ according to classes $c$ in $\gM_p$.\\
            \(\gL_{\mathtt{shift}} \mathrel{+}= \mathtt{MSE}(f_{\theta}(s^c); \hat{p}^c) \). \\ 
        }
    }
    \Return{$\gL_{\mathtt{sync\_proto}} = \alpha\gL_{1} + (1-\alpha)\gL_{2}$}\\
    \caption{Prototypical Exemplar Generation}
    \label{alg:proto-exemplar-2}    
\end{algorithm}

\clearpage
\onecolumn
\section{Experimental Settings}\label{app:settings}
\paragraph{Training Details.} Following existing works \cite{buzzega2020dark}, we adopt ResNet-18 for S-CIFAR-100, and ResNet-50 for S-TinyImageNet, S-ImageNet-1K. For fair comparison, all methods use the same backbone without pretraining and optimizer. The optimizer is performed using Adam optimizer with $\beta_1 = 0.9$ and $\beta_2 = 0.999$. The training epochs vary across datasets: $50$ epochs for CIFAR100, $100$ epochs for S-TinyImageNet, and $100$ for S-ImageNet-1K. We maintain a consistent batch size of 128 across all experiments. Results are averaged over 3 independent runs, and we report the corresponding standard deviation. 

We present the detailed hyper-parameter setting of PEOCL in Table~\ref{tab:hyperparameters}. These hyperparameters are carefully tuned to balance memory efficiency and performance, reflecting the varying complexity of the datasets. The hyperparameter settings of baseline methods are following existing works. For all datasets, we employ minimal data augmentation, consisting of random resized cropping to $224 \times 224$ pixels and random horizontal flipping during training, without any additional augmentation techniques. To prevent overfitting, we set the temperature parameter in the cross-entropy loss to $3$ for all datasets. All experiments were conducted on NVIDIA RTX 4090 GPUs with 24GB memory using PyTorch 3.9.

\begin{table*}[!h]
    \centering
    \setlength{\tabcolsep}{3pt} % Compact column spacing
    \begin{tabular}{c | c | c | c | c | c | c }
    \hline
    {Type} & {Hyperparameters} & {S-CIFAR-100~\cite{dohare2024loss}} & {S-CIFAR-100~\cite{dohare2024loss}} & {S-TinyImageNet~\cite{hou2019learning}} & {S-ImageNet-1K~\cite{dohare2024loss}} \\
    \cline{4-7}
    \hline
    \multirow{5}{*}{\shortstack{Proto\\Network}} 
    & Optimizer & Adam & Adam & Adam & Adam  \\
    & LR & $0.05$ & $0.05$ & $0.05$ & $0.05$  \\
    & Temperature & $0.1$ & $0.1$ & $0.1$ & $0.1$  \\
    & Alignment Coeff. & $0.1$ & $0.1$ & $0.1$ & $0.1$  \\
    & Proto Coeff. & $0.95$ & $0.95$ & $0.95$ & $0.95$  \\
    \hline
    \multirow{5}{*}{\shortstack{Synthetic\\Generator}} 
    & Iterations & $50$ & $50$ & $50$ & $50$  \\
    & Optimizer & AdamW & AdamW & AdamW & AdamW  \\
    & Initial Weight & $10^{-4}$ & $10^{-4}$ & $10^{-4}$ & $10^{-4}$  \\
    & LR & $0.1$ & $0.1$ & $0.1$ & $0.1$  \\
    & LR Scheduler & Cosine Annealing & Cosine Annealing & Cosine Annealing & Cosine Annealing  \\
    \hline
    \end{tabular}
    \vspace{-0.3cm}
    \caption{Hyperparameter settings for different datasets.}
    \label{tab:hyperparameters}
\end{table*}

\paragraph{Evaluation Metrics.} We evaluate our method using two widely adopted metrics in the CL~\cite{buzzega2020dark}. The first metric is the final accuracy, denoted as $A_T$, which measures the model’s overall performance on all tasks after completing the entire training sequence. The second metric is the average accuracy, denoted as $\bar{A}$, which reflects the model’s learning stability over time. It is computed as $\bar{A} = \frac{1}{T}\sum_{t=1}^{T} A_{\tau,t}$, where $T$ is the total number of tasks and $A_{\tau,t}$ is the accuracy on task $t$ after learning task $\tau$. Together, these metrics quantify both the model’s ability to acquire new knowledge and its capability to retain previously learned information, providing a comprehensive evaluation of continual learning performance.

\clearpage
\section{Additional Experiments}
\subsection{Task Incremental Learning}

\begin{table*}[!h]
\centering
\setlength{\tabcolsep}{3pt} % Compact column spacing
\begin{adjustbox}{max width=\textwidth}
{%
\renewcommand{\arraystretch}{1.3}
\begin{tabular}{c | c | c | c c | c c | c c | c c}
\hline
\multirow{2}{*}{Type} & \multirow{2}{*}{Method} & \multirow{2}{*}{Buffer SPCs} & \multicolumn{2}{c|}{S-CIFAR-100~\cite{dohare2024loss}} & \multicolumn{2}{c|}{S-CIFAR-100~\cite{dohare2024loss}} & \multicolumn{2}{c|}{S-TinyImageNet~\cite{hou2019learning}} & \multicolumn{2}{c}{S-ImageNet-1K~\cite{dohare2024loss}} \\
& & & \multicolumn{2}{c|}{$T$=10} & \multicolumn{2}{c|}{$T$=50} & \multicolumn{2}{c|}{$T$=20} & \multicolumn{2}{c}{$T$=100} \\
\cline{4-11}
& & & $\overline{A}$ & $A_T$ & $\overline{A}$ & $A_T$ & $\overline{A}$ & $A_T$ & $\overline{A}$ & $A_T$ \\
\hline
\multirow{5}{*}{\shortstack{Efficient\\Memory}} & CSReL & 20 & $43.86{\scriptstyle\pm1.67}$ & $85.81 {\scriptstyle\pm1.23}$ & $62.97{\scriptstyle\pm1.12}$ & $85.35{\scriptstyle\pm1.71}$ & $25.68{\scriptstyle\pm1.89}$ & $70.49{\scriptstyle\pm1.36}$ & $14.46{\scriptstyle\pm2.26}$ & $\blue{\textbf{67.68}}{\scriptstyle\pm1.62}$ \\
& BCSR & 20 & $50.95 {\scriptstyle\pm1.38}$ & $86.54 {\scriptstyle\pm1.98}$ & $63.48{\scriptstyle\pm0.93}$ & $\blue{\textbf{89.92}}{\scriptstyle\pm2.30}$ & $26.54{\scriptstyle\pm1.26}$ & $71.46{\scriptstyle\pm1.63}$ & $13.97{\scriptstyle\pm2.17}$ & $65.86{\scriptstyle\pm2.49}$ \\
& OCS & 20 & $48.85{\scriptstyle\pm0.88}$ & $85.64{\scriptstyle\pm1.61}$ & $61.24{\scriptstyle\pm1.22}$ & $86.37{\scriptstyle\pm1.49}$ & $23.84{\scriptstyle\pm0.63}$ & $68.46{\scriptstyle\pm1.94}$ & $13.57{\scriptstyle\pm0.42}$ & $66.92{\scriptstyle\pm0.61}$ \\
& PBCS & 20 & $\blue{\textbf{53.36}}{\scriptstyle\pm1.06}$ & $\red{\textbf{91.52}}{\scriptstyle\pm2.03}$ & $\blue{\textbf{64.07}}{\scriptstyle\pm1.13}$ & $\red{\textbf{90.57}}{\scriptstyle\pm1.22}$ & $\blue{\textbf{28.75}}{\scriptstyle\pm0.67}$ & $\red{\textbf{77.09}}{\scriptstyle\pm2.31}$ & $16.14{\scriptstyle\pm1.21}$ & $\red{\textbf{70.46}}{\scriptstyle\pm2.45}$ \\
& OnPro & 20 & $47.51{\scriptstyle\pm1.26}$ & $84.41{\scriptstyle\pm1.83}$ & $60.75{\scriptstyle\pm1.78}$ & $85.03{\scriptstyle\pm2.17}$ & $22.78{\scriptstyle\pm1.18}$ & $70.49{\scriptstyle\pm2.06}$ & $11.28{\scriptstyle\pm0.88}$ & $65.27{\scriptstyle\pm0.33}$ \\
\hline
\multirow{4}{*}{\shortstack{Replay\\based}} & iCaRL & 20 & $43.42{\scriptstyle\pm0.68}$ & $71.46 {\scriptstyle\pm1.61}$ & $58.94 {\scriptstyle\pm0.26}$ & $82.12 {\scriptstyle\pm2.29}$ & $20.34 {\scriptstyle\pm0.84}$ & $67.86 {\scriptstyle\pm1.71}$ & $13.77{\scriptstyle\pm0.89}$ & $60.48{\scriptstyle\pm1.21}$ \\
& DER & 20 & $39.49{\scriptstyle\pm1.52}$ & $80.11{\scriptstyle\pm1.56}$ & $59.36 {\scriptstyle\pm1.13}$ & $81.54 {\scriptstyle\pm0.30}$ & $21.90 {\scriptstyle\pm1.81}$ & $69.23 {\scriptstyle\pm1.16}$ & $13.06{\scriptstyle\pm0.42}$ & $62.87{\scriptstyle\pm0.18}$ \\
& DER++ & 20 & $41.75{\scriptstyle\pm0.16}$ & $81.26{\scriptstyle\pm0.28}$ & $60.77 {\scriptstyle\pm1.71}$ & $85.79{\scriptstyle\pm1.78}$ & $22.06 {\scriptstyle\pm1.36}$ & $70.29 {\scriptstyle\pm1.54}$ & $12.37{\scriptstyle\pm1.36}$ & $64.55{\scriptstyle\pm4.21}$ \\
& C-Flat & 20 & $44.61 {\scriptstyle\pm0.74}$ & $75.13{\scriptstyle\pm0.90}$ & $60.18{\scriptstyle\pm1.25}$ & $84.03{\scriptstyle\pm1.12}$ & $23.97{\scriptstyle\pm0.90}$ & $69.59{\scriptstyle\pm1.27}$ & $13.81{\scriptstyle\pm1.17}$ & $61.46{\scriptstyle\pm1.24}$ \\
\hline
\multirow{3}{*}{\shortstack{Replay\\free}} & LDC & 20 & $37.12{\scriptstyle\pm0.88}$ & $68.92{\scriptstyle\pm0.61}$ & $55.61{\scriptstyle\pm1.26}$ & $75.30{\scriptstyle\pm1.21}$ & $16.61{\scriptstyle\pm0.64}$ & $63.17{\scriptstyle\pm0.74}$ & $12.48{\scriptstyle\pm1.04}$ & $60.87{\scriptstyle\pm1.21}$ \\
& FeCAM & 20 & $36.67{\scriptstyle\pm1.34}$ & $65.14{\scriptstyle\pm1.90}$ & $53.97{\scriptstyle\pm1.25}$ & $69.04{\scriptstyle\pm1.12}$ & $15.47{\scriptstyle\pm0.37}$ & $57.16{\scriptstyle\pm1.23}$ & $11.94{\scriptstyle\pm1.18}$ & $55.78{\scriptstyle\pm1.94}$ \\
& ADC & 20 & $38.35{\scriptstyle\pm0.80}$ & $69.88{\scriptstyle\pm2.04}$ & $56.20{\scriptstyle\pm1.57}$ & $73.58{\scriptstyle\pm1.67}$ & $17.42{\scriptstyle\pm0.58}$ & $64.85{\scriptstyle\pm1.62}$ & $12.91{\scriptstyle\pm1.17}$ & $62.48{\scriptstyle\pm0.98}$ \\
\hline
\multirow{3}{*}{Ours} & ProtoCore & 1 & $39.46{\scriptstyle\pm1.54}$ & $86.13{\scriptstyle\pm0.92}$ & $55.28{\scriptstyle\pm1.13}$ & $87.51{\scriptstyle\pm1.75}$ & $17.42{\scriptstyle\pm1.42}$ & $71.28{\scriptstyle\pm1.94}$ & $15.68{\scriptstyle\pm1.75}$ & $66.38{\scriptstyle\pm1.43}$ \\
& ProtoCore & 5 & $51.12{\scriptstyle\pm1.41}$ & $86.85{\scriptstyle\pm1.57}$ & $63.25{\scriptstyle\pm0.93}$ & $88.49{\scriptstyle\pm1.03}$ & $27.44{\scriptstyle\pm1.24}$ & $\blue{\textbf{72.45}}{\scriptstyle\pm1.77}$ & $\blue{\textbf{18.07}}{\scriptstyle\pm0.89}$ & $66.46{\scriptstyle\pm1.33}$ \\
& ProtoCore & 20 + 1 & $\red{\textbf{58.46}}{\scriptstyle\pm1.58}$ & $\blue{\textbf{87.15}}{\scriptstyle\pm0.91}$ & $\red{\textbf{67.91}}{\scriptstyle\pm1.18}$ & $88.15{\scriptstyle\pm1.59}$ & $\red{\textbf{34.46}}{\scriptstyle\pm1.63}$ & $71.89{\scriptstyle\pm1.46}$ & $\red{\textbf{22.18}}{\scriptstyle\pm1.34}$ & $65.70{\scriptstyle\pm1.83}$ \\
\hline
\end{tabular}
}%
\end{adjustbox}
\vspace{-0.3cm}
\caption{Average ($\overline{A}$) and final ($A_T$) accuracy (\%) comparison on benchmarks with total number of tasks $T$. Results are averaged over 10 runs with mean $\pm$ standard deviation. \red{\textbf{Best}} and \blue{\textbf{Second Best}} results are highlighted.}
\label{tab:extended_results}
\vspace{-0.5cm}
\end{table*}

\subsection{Additional Computation and Memory Usage}\label{app:comp-cost}
\begin{table}[!h]\small
\centering
\caption{Comparison on S-ImageNet-1K (100 tasks) in terms of computation (Hours), GPU and memory usage (GB).}
\label{tab:cost_analysis}
\begin{tabular}{lccc}
\toprule
\textbf{Method} & \textbf{Comp. Time}  & \textbf{GPU Mem.}  & \textbf{HDD Mem.} \\
                & \textbf{(Hours)}      & \textbf{(GB)}      & \textbf{(G)}    
\\
\midrule
CSReL       & 214.61 & 10.04 & 12.08 \\
BCSR        & 287.84 & 24.89 & 12.10 \\
OCS         & 313.44 & 27.55 & 12.94 \\
PBCS        & 391.63 & 18.89 & 12.14 \\
OnPro       & 316.21 & 11.70 & 12.92 \\
iCaRL       & 38.75  & 10.13 & 12.92 \\
DER         & 65.79  & 12.56 & 12.73 \\
DER++       & 83.40  & 23.61 & 12.59 \\
C-Flat      & 105.03 & 19.48 & 12.12 \\
LDC         & 73.20  & 50.50 & 0.47 \\
FeCAM       & 50.28  & 13.45 & 0.00 \\
ADC         & 45.43  & 57.14 & 0.47 \\
ProtoCore   & 252.69 & 30.57 & 1.15 \\
\bottomrule
\end{tabular}
\end{table}

\clearpage
\subsection{Time complexity.} The primary difference between our method and prior work is that we perform synthetic example generation at the end of every task. Unlike generative memory approaches that optimize full network weights, our method optimizes only the learnable quantized latent vector $z$, which has low dimensionality. Consequently, the optimization required for synthetic generation is substantially faster. To quantify this, we measure the wall-clock time consumed by the synthetic generation procedure per class on S-CIFAR-100, S-TinyImageNet, S-ImageNet-1K and report both an empirical per-class time and a simple complexity model.
\begin{table}[!h]
\caption{Analysis on time complexity of the synthetic example generation at the end of every task.}
\begin{center}
\begin{tabular}{ccccc}
\shline
\multirow{2}{*}{\textbf{Method}} & \textbf{S-CIFAR-100 (T=10)} & \textbf{S-ImageNet-1K (T=100)} \\
& \textbf{Seconds/task} & \textbf{Seconds/task} \\ 
\midrule
PBCS & $269.28$ & $11112.84$ \\
CSReL & $151.60$  & $4740.84$ \\
BCSR & $218.16$ & $7368.84$ \\
OCS & $235.31$  & $8304.84$ \\
ProtoCore  & $161.85$ & $6108.84$ \\
\shline 
\end{tabular}
\end{center}
\label{tab:generator-complexity}
\end{table}
\end{document}